\documentclass[10pt,twocolumn,letterpaper]{article}

\usepackage[pagenumbers]{cvpr} %

\definecolor{cvprblue}{rgb}{0.21,0.49,0.74}
\usepackage[pagebackref,breaklinks,colorlinks,allcolors=cvprblue]{hyperref}

\usepackage{multirow}
\usepackage{booktabs}
\usepackage{xcolor}
\usepackage[table]{xcolor}
\usepackage[dvipsnames]{xcolor}
\usepackage{stix}
\usepackage{graphicx}
\usepackage{makecell}

\usepackage{tcolorbox}
\tcbuselibrary{breakable}

\usepackage{xspace}     %
\usepackage{amsmath}    %
\usepackage{amssymb}    %

\usepackage[ruled,vlined]{algorithm2e}
\usepackage{xfp}

\definecolor{uclablue}{RGB}{159, 195, 224}

\definecolor{uclagold}{RGB}{254,180,167}

\definecolor{grayred}{RGB}{232,237,205}

\newcommand{\fire}{\includegraphics[height=0.9em]{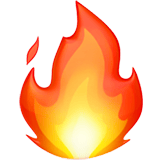}}
\newcommand{\snow}{\includegraphics[height=0.9em]{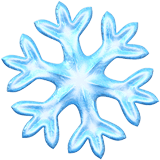}}
\newcommand{\lock}{\includegraphics[height=0.9em]{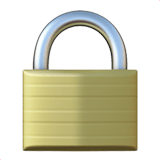}}

\newcommand{\ourname}{RAISE\xspace}

\def\paperID{} %
\def\confName{CVPR}
\def\confYear{2026}

\title{RAISE: Requirement-Adaptive Evolutionary Refinement\\
for Training-Free Text-to-Image Alignment}

\author{Liyao Jiang$^{1}$, Ruichen Chen$^{1}$, Chao Gao$^{2}$, Di Niu$^{1}$\\
$^{1}$Department of ECE, University of Alberta, Canada\quad
$^{2}$Huawei Technologies, Canada\\
{\tt\small \{liyao1,ruichen1,dniu\}@ualberta.ca} \quad
{\tt\small chao.gao4@huawei.com}
}

\begin{document}

\twocolumn[{%
\renewcommand\twocolumn[1][]{#1}%
\maketitle
\centering
\includegraphics[width=0.935\linewidth]{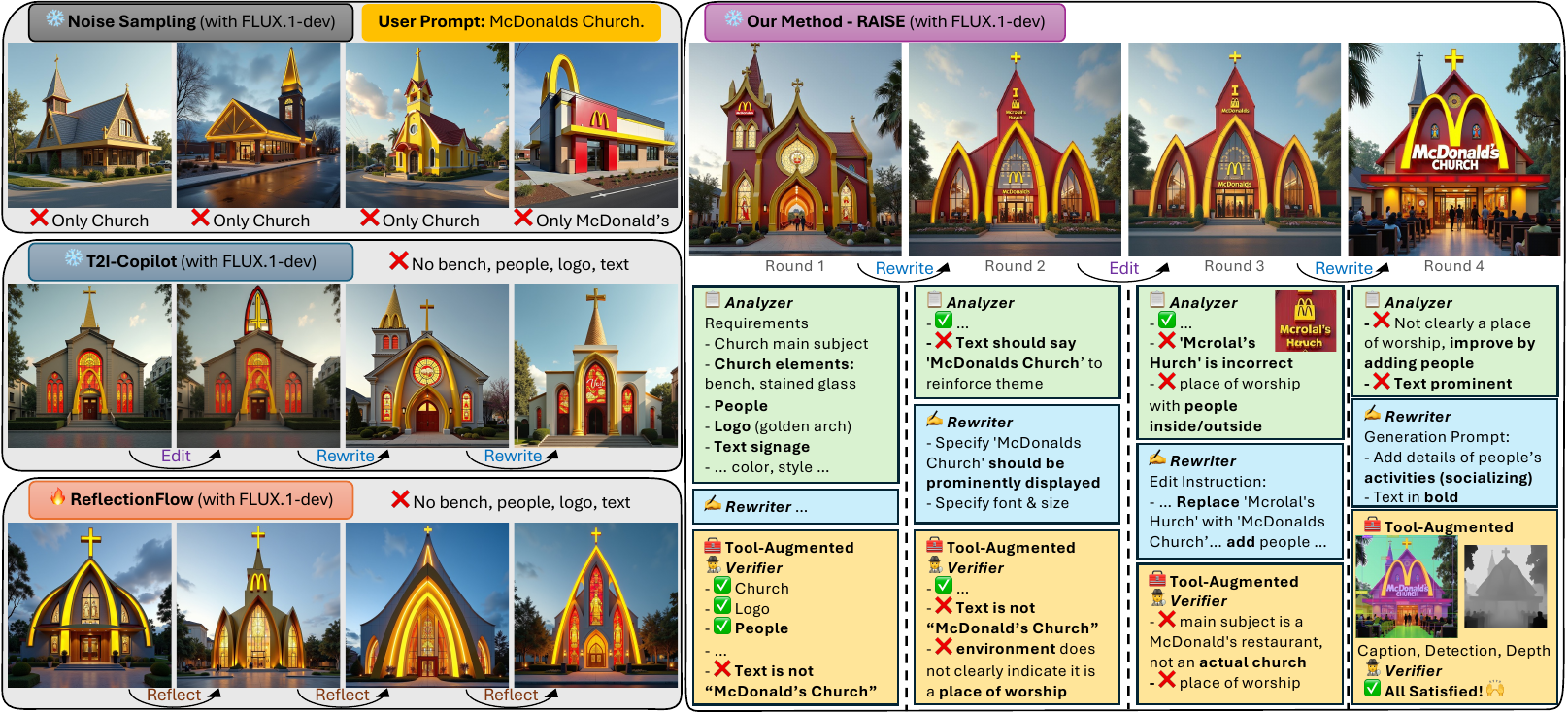}
\captionof{figure}{``McDonald’s Church,'' a challenging prompt.
While other inference-time scaling methods struggle to improve over multiple rounds, \ourname refines T2I alignment using detailed requirement engineering and requirement-driven verification over multiple rounds.}
\vspace{1em}
\label{fig:teaser}
}]

\begin{abstract}
Recent text-to-image (T2I) diffusion models achieve remarkable realism, yet faithful prompt–image alignment remains challenging, particularly for complex prompts with multiple objects, relations, and fine-grained attributes. Existing training-free inference-time scaling methods rely on fixed iteration budgets that cannot adapt to prompt difficulty, while reflection-tuned models require carefully curated reflection datasets and extensive joint fine-tuning of diffusion and vision–language models, often overfitting to reflection paths data and lacking transferability across models.
We introduce RAISE (Requirement-Adaptive Self-Improving Evolution), a training-free, requirement-driven evolutionary framework for adaptive T2I generation. 
RAISE formulates image generation as a requirement-driven adaptive scaling process, evolving a population of candidates at inference time through a diverse set of refinement actions—including prompt rewriting, noise resampling, and instructional editing. 
Each generation is verified against a structured checklist of requirements, enabling the system to dynamically identify unsatisfied items and allocate further computation only where needed. This achieves adaptive test-time scaling that aligns computational effort with semantic query complexity.
On GenEval and DrawBench, RAISE attains state-of-the-art alignment (0.94 overall GenEval) while incurring fewer generated samples (reduced by $30-40\%$) and VLM calls (reduced by $80\%$) than prior scaling and reflection-tuned baselines, demonstrating efficient, generalizable, and model-agnostic multi-round self-improvement. 
Code is available at \url{https://github.com/LiyaoJiang1998/RAISE}.
\end{abstract}
    
\section{Introduction}
\label{sec:introduction_alt}

Recent text-to-image diffusion models (DMs)~\cite{esser2024scaling, flux2024, cai2025hidream, gao2025seedream, lumina2, liu2024playgroundv3improvingtexttoimage, xie2025sana} have achieved remarkable progress in generating photorealistic images from text prompts. 
However, a key challenge is to achieve prompt–image alignment~\cite{wu_towards_2024, hu_ella_2024, jiang2025frap} by generating images that accurately satisfy all semantic and visual requirements described in the prompt, especially for complex prompts which often involve multiple objects, compositional relationships, attribute bindings, and spatial arrangements that need to be coherently represented within a single image~\cite{ghosh2023geneval}. 
While recent unified multimodal models (UMMs)~\cite{xie2024show, chen2025januspro, deng2025emerging, gptimage, wu2025qwenimagetechnicalreport} enhance prompt–image alignment via pre-training on specialized multimodal data, such reliance on large curated datasets remains costly, data-inefficient, and difficult to scale to arbitrary prompts.

Inference-time scaling~\cite{ma2025inference, xie2025sana, khan2025test, chen2025t2icopilot} has emerged as a promising direction to improve text–image alignment by allocating additional computation during T2I generation. 
Noise-level scaling methods~\cite{ma2025inference, xie2025sana} perform random searches guided by scoring functions~\cite{schuhmann2022laion, hessel2021clipscore, xu2023imagereward} to select more optimal initial noise samples~\cite{qi2024not, zhou2025golden}. 
However, noise sampling alone can bring limited improvements to T2I alignment.
Prompt-level scaling methods~\cite{khan2025test, chen2025t2icopilot, zhuo2025reflectionflow} leverage vision–language models (VLMs) to rewrite and refine textual prompts for improved semantic alignment. 
Among them, T2I-Copilot~\cite{chen2025t2icopilot} integrates both noise-level and prompt-level scaling to iteratively refine the prompt.
Yet, these methods typically rely on either a fixed computational budget~\cite{ma2025inference, xie2025sana} or fixed thresholds~\cite{chen2025t2icopilot}, failing to adapt to the varying difficulty of prompts.
Crucially, their refinement strategies demonstrate negligible and even negative impact in subsequent iterations, failing to truly gain from multi-round refinement as demonstrated in Fig. \ref{fig:teaser}.

Other training-based inference-time scaling methods~\cite{li2025reflect, zhuo2025reflectionflow} take a different approach of reflection fine-tuning, %
which is achieved by jointly fine-tuning both the DM and VLM to enable image generations conditioned on previously generated images and their corresponding textual feedback. While these reflection-tuned models significantly improve prompt alignment, they require fine-tuning and a carefully curated large-scale dataset, and thus may often overfit to the reflection paths collected and are not easily transferable to new base models.

To address these challenges, in this paper, we propose Requirement-Adaptive Self-Improving Evolution (RAISE)--a training-free, requirement-driven, and adaptive evolutionary framework for improving text-to-image (T2I) alignment at inference time. Unlike prior works that rely on training-intensive reflection loops to achieve inference-time alignment refinement, RAISE dynamically allocates computational effort based on the semantic complexity and unsatisfied requirements of each prompt. This adaptive design not only enables genuine multi-round self-improvement in image generation without any model retraining, but also achieves substantially higher computational efficiency than existing methods.
Our main contributions are as follows:

\begin{itemize}
   
    \item We formulate T2I alignment as a \textbf{requirement-driven adaptive scaling} process, where an analyzer agent dynamically identifies unsatisfied semantic requirements—including object presence, attributes, and spatial relations—and allocates additional computation only where needed. This enables adaptive inference-time scaling that aligns computational effort with prompt difficulty and automatically converges once all major requirements are satisfied.

    \item  We introduce a \textbf{multi-action evolutionary framework} that concurrently explores complementary refinement strategies, including prompt rewriting, noise resampling, and instructional editing, to enhance both semantic and visual fidelity. This parallel design expands the search space and supports progressive, self-correcting evolution of candidate generations across iterations.

    \item We develop a \textbf{structured verification mechanism} that bridges visual perception and textual reasoning. The verifier agent leverages vision tools—for captioning, detection, and depth estimation—to extract object-level entities, attributes, and spatial relations as evidence for fine-grained requirement checking. This tool-grounded feedback loop enables interpretable, targeted refinement and closes the reasoning–perception gap.
\end{itemize}

Overall, RAISE achieves state-of-the-art prompt–image alignment on GenEval and DrawBench, attaining 0.94 overall GenEval score and 0.885 VQAScore, while requiring substantially fewer generated samples (reduced by $30-40\%$) and VLM calls (reduced by $80\%$) compared to training-based reflection-tuned baselines.
On GenEval, using only FLUX.1-dev~\cite{flux2024} as the base model, our method also beats UMMs like Qwen-Image-RL, BAGEL, and GPT Image 1 which required enormous amounts of pretraining efforts.

\section{Related Work}
\label{sec:related}

\subsection{Training-Free Inference-Time Scaling}

Inference-time scaling improves text–image alignment by allocating extra compute during inference without retraining, unlike model or data scaling which expand model capacity or dataset size.
\citet{ma2025inference} showed that increasing diffusion steps yields diminishing returns compared to resampling initial noise.
Their method performs random search guided by scoring functions~\cite{hessel2021clipscore, xu2023imagereward} to select better initialization latents.  
While simple and training-free, such noise-level scaling relies purely on stochastic variation and lacks semantic reasoning or iterative feedback for improvement.  
SANA-1.5~\cite{xie2025sana} extends this idea by incorporating a VILA-Judge model fine-tuned from VILA-2B~\cite{liu2025nvila} to more reliably assess text–image alignment.
However, both approaches fail to address prompt-image misalignment.  

TIR~\cite{khan2025test} iteratively rewrites prompts using a vision–language model (VLM) to correct mismatches between generated images and intended semantics, leading to improved compositional reasoning.  
However, it operates sequentially and focuses solely on linguistic refinement without leveraging image-level correction.  
T2I-Copilot~\cite{chen2025t2icopilot} combines both noise-level and prompt-level scaling in a training-free agentic loop, where VLM agents evaluates the generated image, reasons about prompt improvement, regenerates images under varied noise, and applies editing to correct inconsistencies.  
Although effective, T2I-Copilot selects a single action per iteration and relies on fixed stopping thresholds, which limits adaptivity and exploration.  
In contrast, \ourname\ adopts a population-based evolutionary framework where multiple refinement strategies including prompt rewriting, noise resampling, and instructional editing operate concurrently.  
This design allows broader exploration and automatically adapts computational effort to prompt complexity through explicit requirement verification.

\subsection{Training-Based Inference-Time Scaling}

Training-based inference-time scaling introduces learned reflection to diffusion models by jointly fine-tuning both the diffusion model (DM) and a VLM to perform in-context reflection.  
Reflect-DiT~\cite{li2025reflect} integrates a context transformer atop the diffusion model and jointly fine-tunes both components, enabling the model to condition on embeddings of previous images and textual feedback.  
The accompanying VLM is also fine-tuned to produce reflection instructions that describe how to improve subsequent generations.  
ReflectionFlow~\cite{zhuo2025reflectionflow} extends this concept by constructing a large-scale reflection dataset (GenRef) containing one million triplets of flawed images, textual reflections, and improved outputs.  
It fine-tunes the DM and VLM jointly to process prior images, feedback, and prompts as unified multimodal tokens.  
While effective, these approaches require costly large-scale joint fine-tuning, making them model-specific and resource-intensive.  
In contrast, \ourname\ achieves reflection-like self-improvement entirely at inference time without any additional training.  
It evolves a diverse population of candidates through parallel mutation strategies and adaptively allocates refinement effort based on requirement satisfaction, achieving stronger alignment with substantially lower computational cost.

\section{Method}
\label{sec:method}

\begin{figure*}[t]
    \centering
    \includegraphics[width=1.0\textwidth]{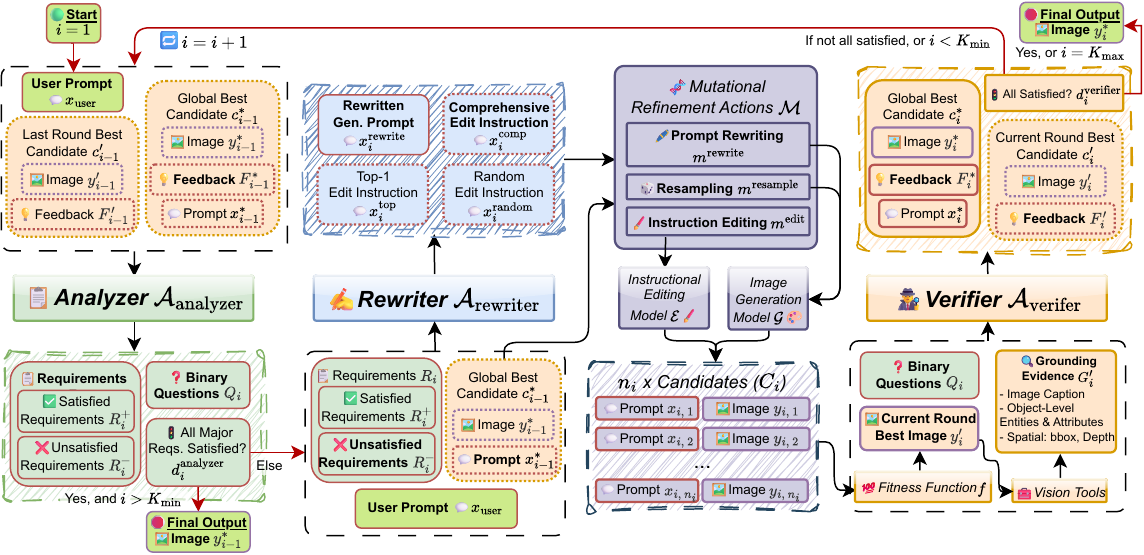}
    \caption{
        \textbf{Framework overview.} 
        \ourname\ employs diverse \textcolor{violet}{mutational refinement actions} concurrently—including prompt rewriting, noise resampling, and instructional editing—to evolve candidates in each round.
        It operates as a multi-agent system composed of an analyzer, rewriter, and verifier: 1) \textcolor{ForestGreen}{Analyzer} performs requirement analysis by extracting a structured and detailed checklist of prompt requirements based on user prompt and previous verification results; 2) \textcolor{Cyan}{Rewriter} refines T2I generation prompts or produces image editing instructions to address unsatisfied requirements; 3) \textcolor{Tan}{Verifier} evaluates generated candidates via structured tool-grounded verification.
    }
    \label{fig:figure_framework}
\end{figure*}

\ourname\ is a training-free, requirement-driven framework that performs adaptive evolutionary scaling for text-to-image (T2I) generation. It progressively improves prompt–image alignment by evolving a population of candidates through iterative cycles of requirement analysis, multi-action mutational refinement, and structured tool-grounded verification. 
By identifying unsatisfied requirements and allocating additional compute only when necessary, \ourname\ achieves \textit{adaptive inference-time scaling}, aligns computational effort with semantic query complexity and converges once all major requirements are satisfied. The overall framework is illustrated in \cref{fig:figure_framework}.

\ourname\ operates as a \textit{multi-agent system} composed of three cooperative agents sharing a common VLM backbone:  
(1) the \textit{analyzer}, which performs requirement analysis by analyzing the user prompt, extracting a structured checklist of requirements, and updating them based on verifier feedback;  
(2) the \textit{rewriter}, which refines generation prompts and produces editing instructions to address unsatisfied requirements; and  
(3) the \textit{verifier}, which evaluates generated candidates against the binary questions corresponding to requirements and provides structured feedback to guide subsequent rounds.  
Detailed agent prompts and the step-by-step algorithm are provided in the supplementary materials.

\cref{sec:requirement} introduces the requirement analysis process and the adaptive scaling mechanism that allocates additional refinement only when needed;  
\cref{sec:refinements} describes the multi-action mutational refinement that generates a diverse candidate population to explore different refinement strategies;  
and \cref{sec:verification} presents the structured tool-grounded verification that leverages vision tools to extract structured grounding evidence to reason over and answer binary verification questions for interpretable and fine-grained assessment.

\subsection{Requirement-Driven Adaptive Scaling}
\label{sec:requirement}

\noindent\textbf{Requirement Analysis.}
A vision–language model (VLM) serves as the \textit{analyzer} agent, identifying the specific requirements to align image generation with the user prompt.

At the beginning of each round $i \in \{1, \dots, K_\text{max}\}$, the analyzer $\mathcal{A}_{\text{analyzer}}$ receives the original prompt $x_{\text{user}}$, the prompt $x^*_{i-1}$ and image $y^*_{i-1}$ of the global best candidate $c^*_{i-1}$, and its verification feedback $\mathcal{F}^*_{i-1}$.  
If the best candidate from the previous round $c'_{i-1}$ differs from the global best $c^*_{i-1}$, i.e., $c'_{i-1}\neq c^*_{i-1}$, the analyzer also takes $\mathcal{F}'_{i-1}$ and $x'_{i-1}$ to maintain context on the evolving requirements.

The analyzer jointly interprets these multimodal inputs and produces structured requirement outputs:
\begin{equation}
\begin{aligned}
O_i^{\text{analyzer}} 
&= \mathcal{A}_{\text{analyzer}}\!\big(x_{\text{user}}, y^*_{i-1}, x^*_{i-1}, \mathcal{F}^*_{i-1}, x'_{i-1}, \mathcal{F}'_{i-1}\big) \\
&= \big(\mathcal{R}_i, \mathcal{R}_i^{+}, \mathcal{R}_i^{-}, Q_i, d_i^{\text{analyzer}}\big),
\end{aligned}
\label{eq:analyzer}
\end{equation}
where $\mathcal{R}_i$ is the complete requirement set (checklist) partitioned into satisfied $\mathcal{R}_i^{+}$ and unsatisfied $\mathcal{R}_i^{-}$ subsets, $Q_i$ is the corresponding binary question set, and $d_i^{\text{analyzer}}$ is the decision variable indicating whether refinement continues.

Each requirement item $r_{i,k} \in \mathcal{R}_i$ represents a verifiable visual condition explicitly or implicitly described in the user prompt, such as object presence, attribute, spatial relation, or composition, paired with a binary question $q_{i,k} \in Q_i$ used in verification to assess satisfaction.

\noindent\textbf{Requirement-Adaptive Scaling.}  
The analyzer $\mathcal{A}_\text{analyzer}$ outputs a decision variable $d_i^{\text{analyzer}} \in \{\text{``continue''}, \text{``end''}\}$ (see \cref{eq:analyzer}).
It ends when all \textit{major requirements}, such as subjects, object count, attributes, spatial relations, or embedded text, are satisfied, and continues otherwise.  
In addition, the verifier $\mathcal{A}_\text{verifier}$ outputs another decision variable $d_i^{\text{verifier}} \in \{\text{True}, \text{False}\}$ (see \cref{eq:verifier}), which is \text{True} only when \textit{all} requirements, both major and minor, are satisfied.  

The iterative process ends when either (1) major requirements are satisfied according to the analyzer, (2) all requirements are satisfied according to the verifier, or (3) the maximum round limit $K_\text{max}$ is reached.  
To ensure sufficient exploration, refinement continues until the minimum round limit $K_\text{min}$.
This scaling mechanism allows \ourname\ to allocate more rounds to unmet requirements and stops automatically once semantic completeness is achieved.

The \textbf{final output image} $y^*$ is produced at round $i$ when $d_i^{\text{analyzer}} = \text{``end''}$ or $d_i^{\text{verifier}} = \text{True}$, or when $K_\text{max}$ rounds are completed, corresponding to the global best candidate $c^*$ with the highest fitness score:
\begin{equation}
c^* = \arg\max_{c_{t,j},\, t \leq i} f(y_{t,j}, x_{\text{user}}),
\quad
y^* = y_{t^*,j^*},
\label{eq:final_output}
\end{equation}
where $f(y_{t,j}, x_{\text{user}})$ denotes the fitness score (see \cref{eq:fitness_selection}), and $(t^*, j^*)$ are the indices of the global maximum.

\subsection{Multi-Action Mutational Refinement}
\label{sec:refinements}

\noindent \textbf{Refinement Actions.}
\ourname\ employs complementary mutational refinement actions to explore diverse directions in the generation and editing spaces concurrently, enabling comprehensive improvement toward unsatisfied requirements.  
It consists of three refinement actions: \textit{1) resampling}, \textit{2) prompt rewriting}, and \textit{3) instructional editing}.

\noindent\textbf{1) Resampling.}
Resampling preserves the original user prompt $x_{\text{user}}$ while exploring alternative visual configurations through stochastic noise sampling:
\begin{equation}
m_{i,j}^{\text{resample}}(c^*_{i-1})
= c_{i,j}
= (\epsilon_{i,j}, x_{\text{user}}, \varnothing),
\quad \epsilon_{i,j} \sim \mathcal{N}(0, I).
\label{eq:mutation_resample}
\end{equation}
By varying only the initial noise, this mutation diversifies spatial layouts, compositions, and object arrangements without altering the prompt semantics.

\noindent\textbf{2) Prompt Rewriting.}
Prompt rewriting refines prompt semantics to address unmet requirements.  
In each round $i$, the \textit{generation rewriter} $\mathcal{A}_\text{rewriter}^\text{gen}$ generates a rewritten prompt $x_i^{\text{rewrite}}$ by applying targeted adjustments derived from the analyzer’s unsatisfied requirement set $\mathcal{R}_i^{-}$:
\begin{equation}
x_i^{\text{rewrite}}
= \mathcal{A}_\text{rewriter}^\text{gen}
\big(x_{\text{user}}, x^*_{i-1}, y^*_{i-1}, \mathcal{R}_i^{+}, \mathcal{R}_i^{-}\big).
\label{eq:rewritten_prompt}
\end{equation}
The rewritten prompt is then paired with multiple independently sampled noises to form diverse candidates:
\begin{equation}
m_{i,j}^{\text{rewrite}}(c^*_{i-1})
= c_{i,j}
= (\epsilon_{i,j}, x_i^{\text{rewrite}}, \varnothing),
\quad \epsilon_{i,j} \sim \mathcal{N}(0, I).
\label{eq:mutation_rewrite}
\end{equation}
This refinement action introduces semantic corrections while maintaining diversity through noise sampling.

\noindent\textbf{3) Instructional Editing.}
Instructional editing operates on the best image $y^*_{i-1}$ to perform refinements guided by textual instructions.  
The \textit{editing rewriter} $\mathcal{A}_\text{rewriter}^\text{edit}$ generates three instruction variants:  
a top edit focusing on the most important unsatisfied requirement,  
a random edit targeting one of the unsatisfied requirements,  
and a comprehensive edit addressing all unsatisfied requirements:
\begin{equation}
\big(x_i^{\text{top}}, x_i^{\text{random}}, x_i^{\text{comp}}\big)
= \mathcal{A}_\text{rewriter}^\text{edit}
\big(x_{\text{user}}, x^*_{i-1}, y^*_{i-1}, \mathcal{R}_i^{+}, \mathcal{R}_i^{-}\big).
\label{eq:editing_prompts}
\end{equation}
Each editing mutation reuses $y^*_{i-1}$ as the reference image and samples a new initial noise $\epsilon_i \sim \mathcal{N}(0, I)$:
\begin{equation}
\begin{aligned}
&m_{i,j}^{\text{edit-variant}}(c^*_{i-1}) = (\epsilon_{i,j}, x_i^{\text{variant}}, y^*_{i-1}), \\
&\text{variant} \in \{\text{top}, \text{random}, \text{comp}\},\ 
\epsilon_{i,j} \sim \mathcal{N}(0, I).
\end{aligned}
\label{eq:mutation_edit_triple}
\end{equation}

The top-edit mutation targets the most critical unsatisfied requirement,  
the random-edit mutation explores alternative correction paths,  
and the comprehensive-edit mutation applies multi-requirement refinements.  
Together, these edits enable focused, diverse, and compounded refinements while preserving global structure and coherence.

\noindent \textbf{Action Execution.}
After the \textit{requirement analysis} and \textit{rewriter} finishes, a set of $n_i$ mutational refinement actions $\mathcal{M}_i=\{m_{i,1}(\cdot), \dots, m_{i,n_i}(\cdot)\}$ are applied to the global best candidate $c^*_{i-1}$, serving as the parent.
Each mutation $m_{i,j}$ generates a new candidate $c_{i,j}$, forming the new population:
\begin{equation}
\begin{aligned}
\mathcal{C}_{i} 
&= \{m_{i,1}(c^*_{i-1}), \dots, m_{i,n_i}(c^*_{i-1})\} 
= \{c_{i,1}, \dots, c_{i,n_i}\}, \\
c_{i,j} &= (\epsilon_{i,j}, x_{i,j}, y'_{i,j}), \quad 
j \in \{1, \ldots, n_i\}.
\end{aligned}
\label{eq:candidate_mutation}
\end{equation}
Here, $\epsilon_{i,j}$ denotes the initial noise, $x_{i,j}$ is the generation or editing prompt, and $y'_{i,j}$ is the optional reference image used for editing-based mutations.
\ourname\ adjusts refinement actions across rounds to balance exploration and refinement. 
Early rounds ($i \le K_{\min}$) use generation-based refinements (\textit{resampling}, \textit{rewriting}) for diverse exploration, while later rounds ($i > K_{\min}$) combine \textit{rewriting} with three editing refinements (\textit{top}, \textit{random}, \textit{comp}) for targeted refinement.

Each candidate $c_{i,j}$ is executed to produce an output image $y_{i,j}$.  
Execution is performed using either the image generation model $\mathcal{G}$ or the image editing model $\mathcal{E}$, depending on whether the candidate includes a reference image $y'_{i,j}$ (i.e., editing-based vs. generation-based mutations):
\begin{equation}
y_{i,j} =
\begin{cases}
\mathcal{G}(\epsilon_{i,j}, x_{i,j}), & \text{if } y'_{i,j} = \varnothing, \\
\mathcal{E}(\epsilon_{i,j}, x_{i,j}, y'_{i,j}), & \text{otherwise},
\end{cases}
\quad \forall j \in \{1, \ldots, n_i\}.
\label{eq:candidate_execution}
\end{equation}
Here, $\mathcal{G}$ denotes the text-to-image generation model that synthesizes images conditioned on the candidate generation prompt $x_{i,j}$ and initial noise $\epsilon_{i,j}$, while $\mathcal{E}$ denotes the instructional image editing model that adjusts a reference image $y'_{i,j}$ using the candidate editing prompt $x_{i,j}$ and initial noise $\epsilon_{i,j}$.

\subsection{Structured Tool-Grounded Verification}
\label{sec:verification}

\noindent\textbf{Fitness Scoring.}  
We score each output image $y_{i,j}$ using a fitness function $f$ that measures alignment with the user prompt $x_{\text{user}}$.  
The highest-scoring candidate in the current round and the global best across all rounds are selected as:
\begin{equation}
\begin{aligned}
s_{i,j} &= f(y_{i,j}, x_{\text{user}}), \\
c'_{i} &= \arg\max_{c_{i,j}} f(y_{i,j}, x_{\text{user}}), \\
c^*_{i} &= \arg\max_{c_{t,j},\, t \leq i} f(y_{t,j}, x_{\text{user}}).
\end{aligned}
\label{eq:fitness_selection}
\end{equation}
Here, $s_{i,j}$ is the fitness of candidate $c_{i,j}$, $c^*_{i}$ the global best up to round $i$, and $c'_{i}$ the round-best selected for verification.

\noindent\textbf{Grounding via Vision Tools.}  
To bridge visual perception and textual reasoning, \ourname\ adopts a tool-grounded verification strategy that leverages vision tools for extracting object-level entities, attributes, and spatial relations into structured textual grounding evidence $G_{i,j}$ to support fine-grained and interpretable verification by VLMs.

\begin{equation}
\begin{aligned}
G_{i,j} = \big(
&c_{i,j}^{\text{det}},\
R_{i,j} = \{ g_k = (l_k, b_k, d_k) \},\
\text{image size } (w, h)
\big),
\end{aligned}
\label{eq:grounding_combined}
\end{equation}
where $c_{i,j}^{\text{det}}$ is the caption describing the overall scene in the image $y_{i,j}$,  
$l_k$ denotes a region label or entity phrase,  
$b_k = [x_{\min}, y_{\min}, x_{\max}, y_{\max}]$ represents the bounding box in xyxy format,  
and $d_k \in [0, 255]$ is the mean depth value estimated within the region.  
The resulting grounding evidence $G_{i,j}$ provides the verifier $\mathcal{A}_\text{verifier}$ with both semantic context and spatial representation for accurate requirement verification.

\noindent\textbf{Structured Binary Checklist Verification.}  
Requirement verification is performed by VLM acting as \textit{verifier} $\mathcal{A}_{\text{verifier}}$, which determines if each requirement $r_{i,k} \in \mathcal{R}_i$ is satisfied by the current round best candidate $c'_{i}$ and its image $y'_{i}$ by answering the corresponding binary question $q_{i,k} \in \mathcal{Q}_i$.

The verifier takes the tool-grounded evidence text $G'_{i}$ derived from $y'_{i}$ together with the binary verification questions set $\mathcal{Q}_i = \{q_{i,1}, \dots, q_{i,n}\}$, and outputs verification results, summary feedback, and a decision variable indicating whether all requirements are satisfied:
\begin{equation}
\begin{aligned}
(\mathcal{F}'_i, d_i^{\text{verifier}}) 
= \big((\mathcal{V}_i, U_i), d_i^{\text{verifier}}\big)
= \mathcal{A}_{\text{verifier}}\!\big(y'_{i}, G'_{i}, \mathcal{Q}_i\big), \\
\mathcal{V}_i 
= \{\,v_{i,k} = (q_{i,k}, a_{i,k}, e_{i,k}) \mid a_{i,k} \!\in\! \{\text{``Yes''}, \text{``No''}\}\,\}.
\end{aligned}
\label{eq:verifier}
\end{equation}
Here, $\mathcal{V}_i$ is the set of verification triplets, where each $v_{i,k}$ contains the binary verification question $q_{i,k}$ for requirement $r_{i,k}$, the verifier’s binary answer $a_{i,k}$ indicating whether it is satisfied, and an explanation $e_{i,k}$ supporting the decision.  
$U_i$ provides a textual summary of satisfied and unsatisfied requirements with suggestions for improving subsequent requirement extraction, while $d_i^{\text{verifier}} \in \{\text{True}, \text{False}\}$ serves as a signal that becomes \text{True} only when \textit{all} requirements are satisfied.  
The overall feedback $\mathcal{F}'_i$ is passed to the analyzer in the next round to guide further evolution.

\begin{table*}[t]
\caption{\textbf{Quantitative comparison on GenEval~\cite{ghosh2023geneval}}. The best and second results are \textbf{bolded} and \underline{underlined}, respectively; category-best methods are also \textbf{bolded}. ``Avg. \#Samples Generated'' and ``Avg. \#Calls VLM'' indicate efficiency.}%
\label{tab:table_geneval}
\centering
\small
\setlength{\tabcolsep}{3.0pt}
\renewcommand{\arraystretch}{0.80}
\resizebox{0.86\textwidth}{!}{
\begin{tabular}{@{}c l | c c | c | c c c c c c@{}}
\toprule
& \multirow{3}{*}{\textbf{Methods}} 
& \textbf{Avg.}
& \textbf{Avg.}
& \multicolumn{7}{c@{}}{\textbf{GenEval Score}} \\ 
\cmidrule(lr){5-11}
& & \textbf{\#Samples} & \textbf{\#Calls} 
& \textbf{Overall} & \textbf{Single} & \textbf{Two} 
& \textbf{Count-} & \multirow{2}{*}{\textbf{Colors}} 
& \textbf{Posit-} & \textbf{Attribute} \\ 
& & \textbf{Generated} & \textbf{VLM} 
& & \textbf{Object} & \textbf{Object} 
& \textbf{ing} & & \textbf{ion} & \textbf{Binding} \\ 
\cmidrule{1-11}

\multicolumn{11}{c}{\cellcolor{lightgray} \textbf{Diffusion Models} (More Models in Supplementary)} \\
\cmidrule{1-11}
& FLUX.1-dev~\cite{flux2024} & 1 & 0 & 0.67  &  0.99 & 0.81 &  0.75 &  0.80 &  0.21 &  0.48 \\
& SD3.5 Large~\cite{esser2024scaling} & 1 & 0 & 0.71 & 0.98 & 0.89 & 0.73 & 0.83 & 0.34 & 0.47 \\
& SANA-1.5 4.8B~\cite{xie2025sana} & 1 & 0 & 0.72 &  0.99 & 0.85 &  0.77 & 0.87  & 0.34  &  0.54  \\
& Lumina-Image 2.0~\cite{lumina2} & 1 & 0 & 0.73 &  0.99 & 0.87 &  0.67 & 0.88  & 0.34  &  0.62  \\
& Playground v3~\cite{liu2024playgroundv3improvingtexttoimage} & 1 & 0 & 0.76  &  0.99  & 0.95 & 0.72  & 0.82  &  0.50 & 0.54 \\
& HiDream-I1-Full~\cite{cai2025hidream} & 1 & 0 & 0.83 & 1.00 & \underline{0.98} & 0.79 & 0.91 & 0.60 & 0.72 \\
& \textbf{Seedream 3.0}~\cite{gao2025seedream} & 1 & 0 & 0.84 & 0.99 & 0.96 & 0.91 & 0.93 & 0.47 & 0.80 \\

\cmidrule{1-11}
\multicolumn{11}{c}{\cellcolor{Tan} \textbf{Unified Multimodal Models}} \\
\cmidrule{1-11}
& Show-o~\cite{xie2024show} & 1 & 1 & 0.53 & 0.95 & 0.52 & 0.49 & 0.82 & 0.11 & 0.28 \\
& Janus-Pro-7B~\cite{chen2025januspro} & 1 & 1 & 0.80  &  0.99  & 0.89 & 0.59  & 0.90  &  0.79 & 0.66 \\
& BAGEL~\cite{deng2025emerging} & 1 & 1 & 0.82  & 0.99 & 0.94 & 0.81 & 0.88 & 0.64 & 0.63 \\
& GPT Image 1 [High]~\cite{gptimage} & 1 & 1 & 0.84 & 0.99 & 0.92 & 0.85 & 0.92 & 0.75 & 0.61 \\
& Qwen-Image~\cite{wu2025qwenimagetechnicalreport} & 1 & 1 & 0.87 & 0.99 & 0.92 & 0.89 & 0.88 & 0.76 & 0.77 \\

& BAGEL + Rewriter~\cite{deng2025emerging} & 1 & 2 & 0.88 & 0.98 & 0.95 & 0.84 & \underline{0.95} & 0.78 & 0.77 \\
& \textbf{Qwen-Image-RL}~\cite{wu2025qwenimagetechnicalreport} & 1 & 1 &\underline{0.91} & 1.00 & 0.95 & \underline{0.93} & 0.92 & \underline{0.87} & \underline{0.83} \\

\cmidrule{1-11}
\multicolumn{11}{c}{\cellcolor{GreenYellow} \textbf{\fire{} Training-Based Inference-Time Scaling (Reflection Tuning)}} \\
\cmidrule{1-11}
& SANA-1.0-1.6B~\cite{xie2024sana} & 1 & 0 & 0.66 &  0.99 & 0.77 &  0.62 & 0.88  & 0.21  &  0.47  \\
& + \fire{} Reflect-DiT~\cite{li2025reflect} & $\leq$20 & $\leq$20 & 0.81 &  0.98 & 0.96 &  0.80 & 0.88  & 0.66  &  0.60  \\ 
\cmidrule{1-11}
\cmidrule{1-11}
& FLUX.1-dev~\cite{flux2024} & 1 & 0 & 0.67  &  0.99 & 0.81 &  0.75 &  0.80 &  0.21 &  0.48 \\
& + \fire{} \textbf{ReflectionFlow}~\cite{zhuo2025reflectionflow} & 32 & 64 & \underline{0.91} & 1.00 & \underline{0.98} & 0.89 & \underline{0.95} & \textbf{0.89} & 0.75 \\

\cmidrule{1-11}
\multicolumn{11}{c}{\cellcolor{Dandelion} \textbf{\snow{} Training-Free Inference-Time Scaling}} \\
\cmidrule{1-11}

& FLUX.1-dev~\cite{flux2024} & 1 & 0 & 0.67  &  0.99 & 0.81 &  0.75 &  0.80 &  0.21 &  0.48 \\
& + \snow{} TIR~\cite{khan2025test} & 4 & 4 & 0.71  & 0.99  & 0.81 & 0.71 & 0.81  & 0.49  & 0.47 \\
& + \snow{} T2I-Copilot~\cite{chen2025t2icopilot} & 11.3 & 22.6 & 0.74 & 0.99 & 0.91 & 0.68 & 0.86 & 0.55 & 0.46 \\
& + \snow{} Noise Scaling~\cite{zhuo2025reflectionflow, ma2025inference} & 32 & 0 & 0.85  & 1.00  & 0.96 & 0.91 & 0.91  & 0.52  & 0.78 \\
& + \snow{} Noise \& Prompt Scaling~\cite{zhuo2025reflectionflow} & 32 & 32 & 0.87 &  0.99 & 0.94 &  0.85 & 0.91  & 0.80 &  0.71  \\

\cmidrule{1-11}
& + \snow{} \textbf{\ourname (Ours)} & 18.6 & 7.3 & \textbf{0.94} & \textbf{1.00} & \textbf{1.00} & \textbf{0.95} & \textbf{0.98} & 0.83 & \textbf{0.87} \\

\bottomrule
\end{tabular}
}
\end{table*}

\section{Experiments}
\label{sec:experiments}

\subsection{Experimental Setup}

\noindent \textbf{Implementation Details.} 
We use FLUX.1-dev~\cite{flux2024} as the image generator and FLUX.1-Kontext-dev~\cite{labs2025flux1kontextflowmatching} as the instructional editor, both with 28 diffusion steps.
Mistral-Small-3.2-24B-Instruct-2506~\cite{mistral} serves as the shared VLM backbone for the agents, coordinated by LangGraph~\cite{langgraph} and served locally via Ollama~\cite{ollama}.
NVILA-Lite-2B-Verifier~\cite{liu2025nvila,xie2025sana} is used as the fitness function.
Tool-grounded verification employs Grounded SAM~2~\cite{ravi2024sam2segmentimages, ren2024grounded, ren2024grounding} and Florence-2~\cite{xiao2023florence} for captioning and detection, and MiDaS~\cite{Ranftl2020, Ranftl2021} for depth estimation.
We set $K_{\max}=4$ and $K_{\min}=2$.
For early rounds ($i \le K_{\min}$), we generate 4 resample and 4 rewrite candidates for broad exploration.
For later rounds ($i > K_{\min}$), we generate 5 rewrite and 3 editing candidates (top, random, comp) for targeted refinement.
Each round produces $n_i = 8$ candidates.

\noindent \textbf{Baselines.} 
We evaluate four categories of text-to-image generation systems.
(1) \textit{Diffusion models:} FLUX.1-dev~\cite{flux2024}, SD3.5 Large~\cite{esser2024scaling}, SANA-1.5 4.8B~\cite{xie2025sana}, Lumina-Image 2.0~\cite{lumina2}, Playground v3~\cite{liu2024playgroundv3improvingtexttoimage}, HiDream-I1-Full~\cite{cai2025hidream}, and Seedream 3.0~\cite{gao2025seedream}.
(2) \textit{Unified multimodal models (UMMs):} Show-o~\cite{xie2024show}, Janus-Pro-7B~\cite{chen2025januspro}, BAGEL~\cite{deng2025emerging}, GPT Image 1 [High]~\cite{gptimage}, and Qwen-Image~\cite{wu2025qwenimagetechnicalreport}.
(3) \textit{Training-based inference-time scaling methods (reflection tuning):} Reflect-DiT~\cite{li2025reflect} and ReflectionFlow~\cite{zhuo2025reflectionflow}.
(4) \textit{Training-free inference-time scaling methods:} Noise Scaling~\cite{ma2025inference, zhuo2025reflectionflow}, Noise \& Prompt Scaling~\cite{ma2025inference}, TIR~\cite{khan2025test}, and T2I-Copilot~\cite{chen2025t2icopilot}.

\begin{figure*}[ht]
    \centering
    \includegraphics[width=0.95\textwidth]{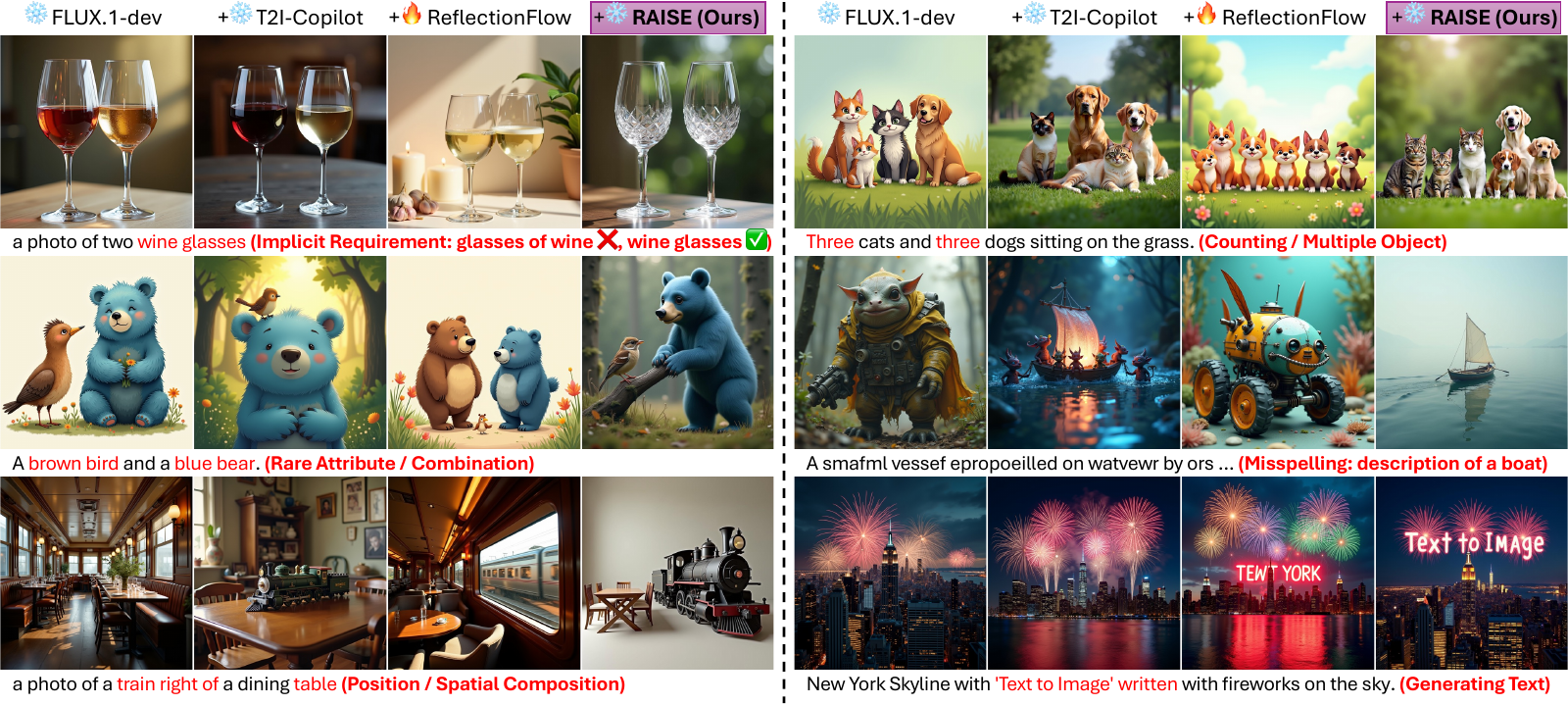}
    \caption{
        \textbf{Visual comparison on GenEval~\cite{ghosh2023geneval} and DrawBench~\cite{saharia2022photorealistic}.}
        \ourname\ improves prompt-image alignment on challenging prompts.
    }
    \label{fig:figure_examples}
\end{figure*}

\begin{figure}[t]
    \centering
    \includegraphics[width=0.93\columnwidth]{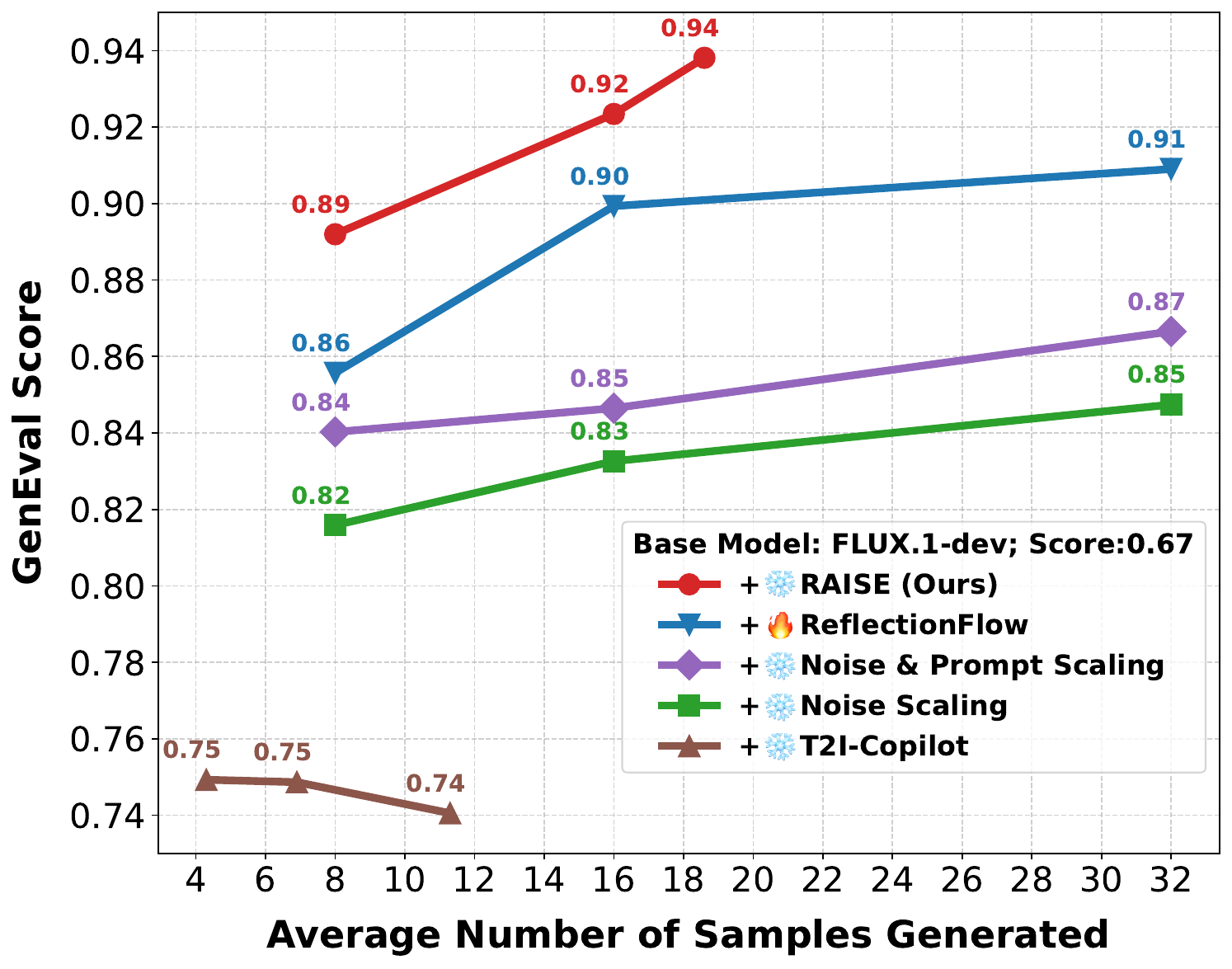}
    \caption{
        \textbf{Pareto frontier and scaling performance.}
        \ourname\ (red) achieves the highest GenEval~\cite{ghosh2023geneval} score with 41.9\% fewer samples (18.6 vs. 32) and 88.6\% fewer VLM calls (7.3 vs. 64).
        Unlike baselines that plateau or fail to improve with additional computation, \ourname\ maintains a strong performance–efficiency Pareto frontier and continues to improve as more samples are generated.
    }
    \label{fig:adaptive-scaling}
\end{figure}

\subsection{Prompt-Image Alignment (GenEval)}

\noindent \textbf{GenEval~\cite{ghosh2023geneval}} is an object-centric evaluation benchmark that measures prompt–image alignment across compositional image properties such as object co-occurrence, position, count, and color.  
It leverages object detection models to assess generation accuracy with strong human correlation and can further verify fine-grained attributes such as object color and spatial relations.  
It contains 553 prompts spanning six compositional categories: \textit{Single Object}, \textit{Two Object}, \textit{Counting}, \textit{Colors}, \textit{Position}, and \textit{Attribute Binding}.

\noindent \textbf{Improved Prompt–Image Alignment.}  
As shown in \Cref{tab:table_geneval}, \ourname\ achieves state-of-the-art result on overall GenEval~\cite{ghosh2023geneval} score (0.94 vs. 0.91), surpassing both training-free inference-time scaling and training-based reflection-tuned approaches, as well as recent proprietary and open-source diffusion models (DMs) and unified multimodal models (UMMs).  
Notably, \ourname\ delivers substantial improvements on challenging categories such as \textit{Counting} (0.95 vs. 0.93), \textit{Colors} (0.98 vs. 0.95), and \textit{Attribute Binding} (0.87 vs. 0.83), where its requirement-driven evolution, multi-action mutational refinement, and tool-grounded verification enhance object presence, attribute binding, and spatial consistency.  
\ourname\ attains 100\% accuracy on the \textit{Two Object} category and 98\% on \textit{Colors}, demonstrating robust alignment even under complex prompts.  

\noindent \textbf{Efficient Adaptive Scaling.}  
\ourname\ achieves superior alignment while achieving better efficiency by dynamically allocating computation.
On average, it generates 41.9\% fewer samples (18.6 vs. 32) and uses 88.6\% fewer VLM calls (7.3 vs. 64) than training-based inference-time scaling approaches on GenEval~\cite{ghosh2023geneval}, without requiring any additional model training.  
This efficiency stems from its requirement-adaptive scaling, directing more computation to semantically complex prompts and converging early on easy ones.
For instance, \ourname on average generates 18.6 samples on GenEval while generating more samples (21.2) on the more complex and reasoning-intensive DrawBench~\cite{saharia2022photorealistic}.

\noindent \textbf{Pareto Frontier and Scaling Analysis.}  
As shown in \cref{fig:adaptive-scaling}, \ourname\ continues to improve as the sampling budget increases, achieving the highest GenEval scores across various budget settings.  
In contrast to reflection-tuned and other training-free inference-time scaling methods that plateau early or fail to improve with additional iterations, \ourname\ maintains a strong performance–efficiency Pareto frontier and achieves steady gains through effective use of additional computation, highlighting the scalability of its requirement-driven, evolutionary multi-action refinement design.

\subsection{Reasoning-Intensive Prompts (DrawBench)}
\begin{table}[t]
\caption{\textbf{Quantitative comparison on the DrawBench~\cite{saharia2022photorealistic}}. The best result in each column is \textbf{bolded}, the second-best is \underline{underlined}. The top-performing setting for each method is also \textbf{bolded}.}
\label{tab:table_drawbench}
\centering
\small
\setlength{\tabcolsep}{1.5pt}
\renewcommand{\arraystretch}{0.80}
\resizebox{0.94\columnwidth}{!}{
\begin{tabular}{@{}l l | c c | c | c c@{}}
\toprule
& \multirow{4}{*}{\textbf{Methods}} 
& \multicolumn{2}{c|}{\textbf{Efficiency}} 
& \multicolumn{1}{c|@{}}{\makecell[c]{\textbf{Prompt--}\\\textbf{Image}\\\textbf{Alignment}}}
& \multicolumn{2}{c@{}}{\makecell[c]{\textbf{Perceptual}\\\textbf{Quality}}} \\ 
\cmidrule(lr){3-7}
& & \textbf{Avg.} & \textbf{Avg.} & \textbf{VQA-} & \textbf{Image-} & \textbf{HPS-} \\
& & \bf \#Samples & \bf \#Calls & \bf Score & \bf Reward & \bf v2 \\
& & \bf Generated & \bf VLM & \bf \cite{lin2024evaluating} & \bf \cite{xu2023imagereward} & \bf \cite{wu2023human} \\
\cmidrule{1-7}

& FLUX.1-dev~\cite{flux2024}                      & 1 & 0 & 0.778 & 1.06 & 0.298 \\
\cmidrule{1-7}
& + \snow{} \textbf{T2I-Copilot}~\cite{chen2025t2icopilot}         & \textbf{3.9} & \textbf{7.9} & 0.822 & 0.97 & 0.300 \\
& + \snow{} T2I-Copilot~\cite{chen2025t2icopilot}         & 6.6 & 13.1 & 0.820 & 0.96 & 0.299 \\
& + \snow{} T2I-Copilot~\cite{chen2025t2icopilot}         & 11.2 & 22.3 & 0.820 & 0.94 & 0.298 \\
\cmidrule{1-7}
& + \fire{} ReflectionFlow~\cite{zhuo2025reflectionflow}  & 8 & 16 & 0.839 & 1.08 & 0.301 \\
& + \fire{} \textbf{ReflectionFlow}~\cite{zhuo2025reflectionflow}  & \textbf{16} & \textbf{32} & 0.844 & \underline{1.13} & 0.302 \\
& + \fire{} ReflectionFlow~\cite{zhuo2025reflectionflow}  & 32 & 64 & 0.844 & 1.10 & 0.302 \\
\cmidrule{1-7}
& + \snow{} \ourname (1 round)                            & 8 & 3 & 0.868 & \underline{1.13} & \underline{0.304} \\
& + \snow{} \ourname (2 rounds)                           & 16 & 6 & \underline{0.876} & \textbf{1.15} & \textbf{0.305} \\
& + \snow{} \textbf{\ourname} ($\leq$4 rounds)            & \textbf{21.2} & \textbf{8.6} & \textbf{0.885} & \textbf{1.15} & \textbf{0.305} \\
\bottomrule
\end{tabular}
}
\end{table}

DrawBench~\cite{saharia2022photorealistic} is a reasoning-intensive benchmark consisting of 200 open-ended prompts evaluating compositionality, counting, spatial relations, rare concepts, and robustness to complex or creative descriptions that fall outside typical training distributions.
We use DrawBench to evaluate reasoning-oriented scenarios such as conflicting, descriptive, misspelled, and text-in-image prompts, assessing prompt–image alignment via VQAScore~\cite{lin2024evaluating} and perceptual quality via ImageReward~\cite{xu2023imagereward} and HPSv2~\cite{wu2023human}.  

\noindent \textbf{Enhanced Reasoning with Stable Perceptual Quality.}
On DrawBench (\cref{tab:table_drawbench}), \ourname\ outperforms both training-free and reflection-tuned baselines across all metrics. It achieves significantly higher prompt–image alignment (0.885 vs. 0.844 VQAScore) with 33.8\% fewer generated samples (21.2 vs. 32) and 86.6\% fewer VLM calls (8.6 vs. 64).
Moreover, \ourname\ not only continues to improve alignment as more samples are generated, but also maintains stable perceptual quality in ImageReward and HPS-v2, unlike other methods that plateau in alignment and degrade in perceptual quality due to error accumulation.
These results show that \ourname substantially enhances alignment and reasoning, while delivering stable performance and even moderate gains in visual realism under complex reasoning conditions.

\subsection{Visual Comparisons}
\noindent \textbf{Requirement-Adaptive Evolutionary Refinement.}  
As shown in \cref{fig:teaser}, \ourname\ improves alignment by identifying explicit and implicit requirements and iteratively correcting unsatisfied ones, including text accuracy, implicit mood and environment, and semantically related elements or actions. Through rewrite, resample, and edit mutations guided by structured tool-grounded verification, it refines missing details such as logo, signage, people, and church attributes, and converges once all requirements are satisfied.  
In contrast, noise sampling, T2I-Copilot, and ReflectionFlow fail to infer implicit requirements, often producing incomplete scenes lacking nuanced details such as people, correct text signage, or key church elements.
See the supplementary materials for a visualization of the evolutionary search path.

\noindent \textbf{Alignment in Challenging Prompts.}
As shown in \cref{fig:figure_examples}, \ourname\ achieves stronger alignment and better perceptual quality than FLUX.1-dev, T2I-Copilot, and ReflectionFlow on challenging GenEval~\cite{ghosh2023geneval} and DrawBench~\cite{saharia2022photorealistic} prompts. It correctly handles implicit requirements (e.g., ``wine glasses’’), multi-object counting, rare attribute combinations, misspelled descriptive prompt, spatial composition, and text-in-image generation, where prior methods frequently misinterpret or omit key details.

\begin{table}[t]
\caption{\textbf{Ablation Studies on the GenEval~\cite{ghosh2023geneval} benchmark.} The best result in each column is \textbf{bolded}, the second-best is \underline{underlined}.}
\label{tab:table_ablation}
\centering
\small
\setlength{\tabcolsep}{3.0pt}
\renewcommand{\arraystretch}{0.8}
\resizebox{0.99\columnwidth}{!}{
\begin{tabular}{@{}l l | c | c c c@{}}
\toprule
& \multirow{2}{*}{\textbf{Methods}} 
& \multirow{2}{*}{\textbf{Overall}} 
& \multirow{2}{*}{\textbf{Colors}} 
& \multirow{2}{*}{\textbf{Position}}
& \textbf{Attribute} \\
& & & & & \bf Binding \\

\cmidrule{1-6}
& FLUX.1-dev~\cite{flux2024} & 0.67 &  0.80 &  0.21 &  0.48 \\
\cmidrule{1-6}
& + \snow{} \ourname (1 round)                 & 0.89 & 0.95 & 0.69 & 0.79 \\
& + \snow{} \ourname (2 rounds)                & 0.92 & \underline{0.97} & 0.78 & 0.83 \\
& + \snow{} \ourname ($\leq$ 3 rounds)         & \underline{0.93} & \textbf{0.98} & \underline{0.82} & \underline{0.86} \\
\cmidrule{1-6}
& + \snow{} \textbf{\ourname} ($\leq$4 rounds) & \textbf{0.94} & \textbf{0.98} & \textbf{0.83} & \textbf{0.87} \\
& + \snow{}$\rightarrow$ w/o Vision Tools        & \underline{0.93} & \underline{0.97} & 0.81 & 0.84 \\
& + \snow{}$\rightarrow$ w/o Editing          & \underline{0.93} & \textbf{0.98} & \textbf{0.83} & 0.83 \\

\bottomrule
\end{tabular}
}
\end{table}

\subsection{Ablation Studies}

We conduct ablation studies on the GenEval~\cite{ghosh2023geneval} benchmark by selectively removing the vision tool grounding and instructional editing modules from \ourname.  
As shown in \Cref{tab:table_ablation}, the full framework achieves the highest overall score (0.94) across all compositional categories, validating the synergy between tool-grounded verification and multi-action refinement.  
Removing vision tools (\textit{w/o Vision Tools}) results in a noticeable drop in \textit{Attribute Binding} and \textit{Colors}, indicating that structured visual evidence is critical for fine-grained attribute alignment.  
Similarly, disabling instructional editing (\textit{w/o Editing}) slightly lowers \textit{Attribute Binding} and overall performance, showing that instructional editing complements prompt rewriting for targeted correction.  
These results confirm that both tool grounding and editing are indispensable for achieving prompt–image alignment.
In addition, experiments in the supplementary materials show that \ourname is model-agnostic and delivers consistent gains across different base DMs and VLMs.

\section{Conclusion}
\label{sec:conclusion}

We present \ourname, a training-free, requirement-driven evolutionary framework, which outperforms training-based inference-time scaling method with 40\% fewer generated samples. The observed gains in \ourname are derived from its multi-round refinement capability. Furthermore, our method proposes a novel way to truly achieve effective multi-round refinement.
Our work offers significant contributions to the advancement of inference-time scaling methods, especially when addressing challenging and nuanced prompts that mandate multi-round refinement.

\clearpage
{
    \small
    \bibliographystyle{ieeenat_fullname}
    \bibliography{main}
}

\clearpage

\twocolumn[{%
\renewcommand\twocolumn[1][]{#1}%
\maketitlesupplementary
\begin{center}
\captionof{table}{
\textbf{Evaluation with different base DMs (FLUX.1-dev~\cite{flux2024}, FLUX.1-schnell~\cite{flux2024}, SANA-1.5 4.8B~\cite{xie2025sana}) on GenEval~\cite{ghosh2023geneval}}. The best and second best results are \textbf{bolded} and \underline{underlined}. ``Avg. \#Samples Generated'' and ``Avg. \#Calls VLM'' indicate efficiency.}
\label{tab:table_base_dm}
\centering
\small
\setlength{\tabcolsep}{3.0pt}
\renewcommand{\arraystretch}{1.0}
\resizebox{0.85\textwidth}{!}{
\begin{tabular}{@{}c l c | c c | c | c c c c c c@{}}
\toprule
& \multirow{3}{*}{\textbf{Methods}}
& \textbf{Diffusion}
& \textbf{Avg.}
& \textbf{Avg.}
& \multicolumn{7}{c@{}}{\textbf{GenEval Score}} \\ 
\cmidrule(lr){6-12}
& 
& \textbf{Model's}
& \textbf{\#Samples} 
& \textbf{\#Calls} 
& \textbf{Overall} & \textbf{Single} & \textbf{Two} 
& \textbf{Count-} & \multirow{2}{*}{\textbf{Colors}} 
& \textbf{Posit-} & \textbf{Attribute} \\ 
&  
& \textbf{Steps / Size}
& \textbf{Generated} 
& \textbf{VLM} 
& & \textbf{Object} & \textbf{Object} 
& \textbf{ing} & & \textbf{ion} & \textbf{Binding} \\ 
\cmidrule{1-12}

& FLUX.1-dev~\cite{flux2024} & 28 / 12B & 1 & 0 & 0.67  & \underline{0.99} & 0.81 & 0.75 & 0.80 & 0.21 & 0.48 \\
& + \snow{} T2I-Copilot~\cite{chen2025t2icopilot} & 28 / 12B & 11.3 & 22.6 & 0.74 & \underline{0.99} & 0.91 & 0.68 & 0.86 & 0.55 & 0.46 \\
& + \fire{} ReflectionFlow~\cite{zhuo2025reflectionflow} & 28 / 12B & 32 & 64 & 0.91 & \textbf{1.00} & \underline{0.98} & 0.89 & 0.95 & \underline{0.89} & 0.75 \\
& + \snow{} \ourname (Ours) & 28 / 12B & 18.6 & 7.3 & \textbf{0.94} & \textbf{1.00} & \textbf{1.00} & \textbf{0.95} & \textbf{0.98} & 0.83 & \textbf{0.87} \\
\cmidrule{1-12}

& FLUX.1-schnell~\cite{flux2024} & 4 / 12B
& 1 & 0 & 0.66  & \textbf{1.00} & 0.87 & 0.59 & 0.76 & 0.29 & 0.45 \\
& + \snow{} \ourname (Ours) & 4 / 12B
& 18.7 & 7.3 & \underline{0.93} & \textbf{1.00} & \textbf{1.00} & \underline{0.93} & \textbf{0.98} & 0.83 & \underline{0.85} \\
\cmidrule{1-12}

& SANA-1.5 4.8B~\cite{xie2025sana} & 20 / 4.8B
& 1 & 0 & 0.72 & \underline{0.99} & 0.85 & 0.77 & 0.87 & 0.34 & 0.54 \\
& + \snow{} \ourname (Ours) & 20 / 4.8B
& 19.4 & 7.7 & \underline{0.93} & \textbf{1.00} & \textbf{1.00} & 0.91 & \underline{0.96} & \textbf{0.92} & 0.81 \\

\bottomrule
\end{tabular}
}
\end{center}

\begin{center}
\captionof{table}{
\textbf{Evaluation of \ourname with different base VLMs~\cite{gemma3,qwen3vl,mistral} on GenEval~\cite{ghosh2023geneval}}. The best and second best results are \textbf{bolded} and \underline{underlined}. ``Avg. \#Samples Generated'' and ``Avg. \#Calls VLM'' indicate efficiency.
\lock{} denotes proprietary models, \fire{} denotes fine-tuned open-source models, and \snow{} denotes frozen open-source models that do not require any additional fine-tuning.}
\label{tab:table_base_vlm}
\centering
\small
\setlength{\tabcolsep}{3.0pt}
\renewcommand{\arraystretch}{1.0}
\resizebox{1.0\textwidth}{!}{
\begin{tabular}{@{}c l | c c | c | c c c c c c@{}}
\toprule
& \multirow{3}{*}{\textbf{Methods}}
& \textbf{Avg.}
& \textbf{Avg.}
& \multicolumn{7}{c@{}}{\textbf{GenEval Score}} \\ 
\cmidrule(lr){5-11}
& 
& \textbf{\#Samples} 
& \textbf{\#Calls} 
& \textbf{Overall} & \textbf{Single} & \textbf{Two} 
& \textbf{Count-} & \multirow{2}{*}{\textbf{Colors}} 
& \textbf{Posit-} & \textbf{Attribute} \\ 
&  
& \textbf{Generated} 
& \textbf{VLM} 
& & \textbf{Object} & \textbf{Object} 
& \textbf{ing} & & \textbf{ion} & \textbf{Binding} \\ 
\cmidrule{1-11}

& FLUX.1-dev~\cite{flux2024} & 1 & 0 & 0.67  & \underline{0.99} & 0.81 & 0.75 & 0.80 & 0.21 & 0.48 \\
\cmidrule{1-11}

& T2I-Copilot~\cite{chen2025t2icopilot} & \multirow{2}{*}{11.3} & \multirow{2}{*}{22.6} & \multirow{2}{*}{0.74} & \multirow{2}{*}{\underline{0.99}} & \multirow{2}{*}{0.91} & \multirow{2}{*}{0.68} & \multirow{2}{*}{0.86} & \multirow{2}{*}{0.55} & \multirow{2}{*}{0.46} \\
& (\snow{} FLUX.1-dev~\cite{flux2024} + \snow{} Mistral-Small-3.2-24B~\cite{mistral}) &  &  &  &  &  &  &  &  & \\
\cmidrule{1-11}

& ReflectionFlow~\cite{zhuo2025reflectionflow} & \multirow{2}{*}{32} & \multirow{2}{*}{64} & \multirow{2}{*}{0.91} & \multirow{2}{*}{\textbf{1.00}} & \multirow{2}{*}{0.98} & \multirow{2}{*}{0.89} & \multirow{2}{*}{0.95} & \multirow{2}{*}{\textbf{0.89}} & \multirow{2}{*}{0.75} \\
& (\fire{} FLUX.1-dev~\cite{flux2024} + \lock{} GPT-4o~\cite{hurst2024gpt} + \fire{} Qwen2.5-VL-7B~\cite{Qwen2.5-VL}) &  &  & & &  & & & & \\
\cmidrule{1-11}

& \ourname (\snow{} FLUX.1-dev~\cite{flux2024} + \snow{} gemma-3-27b-it~\cite{gemma3})
& 18.4 & 7.2 & 0.92 & \textbf{1.00} & \underline{0.99} & \underline{0.93} & \underline{0.97} & 0.78 & \underline{0.86} \\

& \ourname (\snow{} FLUX.1-dev~\cite{flux2024} + \snow{} Qwen3-VL-32B-Instruct~\cite{qwen3vl})
& 20.3 & 8.1 & \underline{0.93} & \textbf{1.00} & \textbf{1.00} & 0.91 & 0.94 & \textbf{0.89} & 0.83 \\

& \ourname (\snow{} FLUX.1-dev~\cite{flux2024} + \snow{} Mistral-Small-3.2-24B~\cite{mistral})
& 18.6 & 7.3 & \textbf{0.94} & \textbf{1.00} & \textbf{1.00} & \textbf{0.95} & \textbf{0.98} & \underline{0.83} & \textbf{0.87} \\

\bottomrule
\end{tabular}
}
\end{center}

}]

\begin{table*}[ht]
\caption{\textbf{Efficiency comparison on GenEval~\cite{ghosh2023geneval}.} 
\ourname\ consistently achieves the highest GenEval score across budgets (Max \#Samples = 8, 16, 32). 
At 32 samples, it requires \textit{41.9\% fewer samples generated} and \textit{88.6\% fewer VLM calls} on average than the second-best method.
}
\label{tab:table_efficiency}
\centering
\small
\setlength{\tabcolsep}{2.5pt}
\renewcommand{\arraystretch}{0.8}
\resizebox{0.9\textwidth}{!}{
\begin{tabular}{@{}l l | c | c | c | c | c }
\toprule
& \textbf{Methods} 
& \textbf{GenEval Score} 
& Max. \#Samples Allowed 
& \textbf{Avg. \#Samples Generated} 
& Max. \#Calls VLM 
& \textbf{Avg. \#Calls VLM} \\
\cmidrule{1-7}
& FLUX.1-dev~\cite{flux2024}                        & 0.67 & 1 & 1 & 0 & 0 \\
\cmidrule{1-7}
& + \snow{} T2I-Copilot~\cite{chen2025t2icopilot}                        & 0.75 & 8 & \textbf{4.3} & 17 & 8.6 \\
& + \fire{} ReflectionFlow~\cite{zhuo2025reflectionflow}                 & 0.86 & 8 & 8 & 16 & 16 \\
& + \snow{} \textbf{\ourname} (1 round)                                  & \textbf{0.89} & 8 & 8 & 3 & \textbf{3} \\
\cmidrule{1-7}
& + \snow{} T2I-Copilot~\cite{chen2025t2icopilot}                        & 0.75 & 16 & \textbf{6.9}  & 33 & 13.8 \\
& + \fire{} ReflectionFlow~\cite{zhuo2025reflectionflow}                 & 0.90 & 16 & 16 & 32 & 32 \\
& + \snow{} \textbf{\ourname} (2 rounds)                                 & \textbf{0.92} & 16 & 16 & 6 & \textbf{6} \\
\cmidrule{1-7}
& + \snow{} T2I-Copilot~\cite{chen2025t2icopilot}                        & 0.74 & 32 & \textbf{11.3} & 65 & 22.6 \\
& + \fire{} ReflectionFlow~\cite{zhuo2025reflectionflow}                 & 0.91 & 32 & 32 & 64 & 64 \\
& + \snow{} \textbf{\ourname} ($\leq$ 4 rounds)                          & \textbf{0.94} & 32 & 18.6 & 14 & \textbf{7.3} \\

\bottomrule
\end{tabular}
}
\end{table*}

\section{Additional Results}

\subsection{\ourname with Different Base Models}
\label{sec:base_models}
\ourname is a training-free and plug-and-play framework that requires no additional tuning of either the diffusion model (DM) or the vision–language model (VLM), allowing it to be seamlessly applied to new DM base models and new VLM base models. As shown in this section, \ourname\ is model-agnostic and delivers consistent gains across both types of base models, demonstrating efficient and generalizable multi-round self-improvement.

\noindent \textbf{Different Base Diffusion Models (DMs).}
\Cref{tab:table_base_dm} reports the results of \ourname\ when paired with a range of DMs. Across all tested DMs, which vary in model size, number of diffusion steps, and generation quality, \ourname\ consistently improves prompt–image alignment and achieves high overall GenEval scores of 0.93–0.94. These results indicate that \ourname\ is robust to the choice of diffusion model and operates in a plug-and-play manner without any DM fine-tuning.

\noindent \textbf{Different Base Vision-Language Models (VLMs).}
\Cref{tab:table_base_vlm} summarizes the performance of \ourname\ when combined with different VLM backbones. Across all tested VLMs, which vary in family and parameter size, \ourname\ delivers consistent improvements in alignment and strong overall GenEval scores of 0.92-0.94. This shows that \ourname\ adapts effectively to different VLM reasoning capabilities, despite differences in architecture, scale, and training strategy. It requires no VLM fine-tuning or proprietary models, highlighting its model-agnostic plug-and-play nature.

\subsection{Details on Efficiency and Adaptive-Scaling}
\label{sec:efficiency}

As shown in \Cref{tab:table_efficiency} and \cref{fig:adaptive-scaling}, \ourname\ continues to improve as the sampling budget increases (max number of samples allowed: 8, 16, 32), achieving the highest GenEval~\cite{ghosh2023geneval} scores across all budget settings.
\ourname achieves superior alignment while achieving better efficiency by dynamically allocating computation.
For example, with a max budget of 32 samples, it generates 41.9\% fewer samples (18.6 vs. 32) and requires 88.6\% fewer VLM calls (7.3 vs. 64) than training-based inference-time scaling approaches on GenEval~\cite{ghosh2023geneval}, while requiring no additional model training.

In contrast to reflection-tuned and other training-free inference-time scaling methods that plateau early or fail to improve with additional samples, \ourname\ maintains a strong performance–efficiency Pareto frontier and achieves steady gains through effective use of additional computation, highlighting the scalability of its requirement-driven, evolutionary multi-action refinement design.

The efficiency of \ourname stems from its requirement-adaptive scaling, directing more computation to semantically complex prompts and converging early on easy ones.
For instance, \ourname on average generates 18.6 samples on GenEval (see \Cref{tab:table_efficiency}) while generating more samples (21.2) on the more complex and reasoning-intensive DrawBench~\cite{saharia2022photorealistic} (see \Cref{tab:table_drawbench}).
In addition, as shown in the last row of \Cref{tab:table_geneval_full}, \ourname\ adaptively invests additional samples in challenging categories of GenEval~\cite{ghosh2023geneval} such as \textit{Colors}, \textit{Position}, and \textit{Attribute Binding}, while minimizing redundant refinements in other categories.

\begin{table*}[ht]
\caption{\textbf{Quantitative comparison of \ourname on the GenEval benchmark~\cite{ghosh2023geneval}} against diffusion models, unified multimodal models, training-free inference-time scaling, and training-based reflection tuning methods. The best and second results are \textbf{bolded} and \underline{underlined}, respectively; category-best methods are also \textbf{bolded}. ``Avg. \#Samples Generated'' and ``Avg. \#Calls VLM'' indicate efficiency, and the last row shows \ourname’s adaptive allocation of more computation to harder categories such as ``Colors'', ``Position'', and ``Attribute Binding''.}
\label{tab:table_geneval_full}
\centering
\small
\setlength{\tabcolsep}{3.5pt}
\renewcommand{\arraystretch}{0.9}
\resizebox{1.0\textwidth}{!}{
\begin{tabular}{@{}c l | c c | c | c c c c c c@{}}
\toprule
& \multirow{3}{*}{\textbf{Methods}} 
& \textbf{Avg.}
& \textbf{Avg.}
& \multicolumn{7}{c@{}}{\textbf{GenEval Score}} \\ 
\cmidrule(lr){5-11}
& & \textbf{\#Samples} & \textbf{\#Calls} 
& \textbf{Overall} & \textbf{Single} & \textbf{Two} 
& \textbf{Count-} & \multirow{2}{*}{\textbf{Colors}} 
& \textbf{Posit-} & \textbf{Attribute} \\ 
& & \textbf{Generated} & \textbf{VLM} 
& & \textbf{Object} & \textbf{Object} 
& \textbf{ing} & & \textbf{ion} & \textbf{Binding} \\ 
\cmidrule{1-11}

\multicolumn{11}{c}{\cellcolor{lightgray} \textbf{Diffusion Models}} \\
\cmidrule{1-11}
& PixArt-$\alpha$~\cite{chen2024pixartalpha} & 1 & 0 & 0.48 & 0.98 & 0.50 & 0.44 & 0.80 & 0.08 & 0.07 \\
& Emu3-Gen~\cite{wang2024emu3} & 1 & 0 & 0.54 & 0.98 & 0.71 & 0.34 & 0.81 & 0.17 & 0.21 \\
& SDXL~\cite{podell2023sdxl} & 1 & 0 & 0.55  &  0.98 & 0.74 &  0.39 &  0.85 &  0.15 &  0.23 \\
& SD3 Medium~\cite{esser2024scaling} & 1 & 0 & 0.62 & 0.98 & 0.74 & 0.63 & 0.67 & 0.34 & 0.36 \\
& JanusFlow~\cite{ma2025janusflow} & 1 & 0 & 0.63 & 0.97 & 0.59 & 0.45 & 0.83 & 0.53 & 0.42 \\
& DALLE 3~\cite{betker2023improving} & 1 & 0 & 0.67  &  0.96 & 0.87 &  0.47 &  0.83 &  0.43 &  0.45 \\
& FLUX.1-dev~\cite{flux2024} & 1 & 0 & 0.67  &  0.99 & 0.81 &  0.75 &  0.80 &  0.21 &  0.48 \\
& SD3.5 Large~\cite{esser2024scaling} & 1 & 0 & 0.71 & 0.98 & 0.89 & 0.73 & 0.83 & 0.34 & 0.47 \\
& SANA-1.5 4.8B~\cite{xie2025sana} & 1 & 0 & 0.72 &  0.99 & 0.85 &  0.77 & 0.87  & 0.34  &  0.54  \\
& Lumina-Image 2.0~\cite{lumina2} & 1 & 0 & 0.73 &  0.99 & 0.87 &  0.67 & 0.88  & 0.34  &  0.62  \\
& Playground v3~\cite{liu2024playgroundv3improvingtexttoimage} & 1 & 0 & 0.76  &  0.99  & 0.95 & 0.72  & 0.82  &  0.50 & 0.54 \\
& HiDream-I1-Full~\cite{cai2025hidream} & 1 & 0 & 0.83 & 1.00 & \underline{0.98} & 0.79 & 0.91 & 0.60 & 0.72 \\
& \textbf{Seedream 3.0}~\cite{gao2025seedream} & 1 & 0 & 0.84 & 0.99 & 0.96 & 0.91 & 0.93 & 0.47 & 0.80 \\

\cmidrule{1-11}
\multicolumn{11}{c}{\cellcolor{Tan} \textbf{Unified Multimodal Models}} \\
\cmidrule{1-11}
& Show-o~\cite{xie2024show} & 1 & 1 & 0.53 & 0.95 & 0.52 & 0.49 & 0.82 & 0.11 & 0.28 \\
& Janus-Pro-7B~\cite{chen2025januspro} & 1 & 1 & 0.80  &  0.99  & 0.89 & 0.59  & 0.90  &  0.79 & 0.66 \\
& BAGEL~\cite{deng2025emerging} & 1 & 1 & 0.82  & 0.99 & 0.94 & 0.81 & 0.88 & 0.64 & 0.63 \\
& GPT Image 1 [High]~\cite{gptimage} & 1 & 1 & 0.84 & 0.99 & 0.92 & 0.85 & 0.92 & 0.75 & 0.61 \\
& Qwen-Image~\cite{wu2025qwenimagetechnicalreport} & 1 & 1 & 0.87 & 0.99 & 0.92 & 0.89 & 0.88 & 0.76 & 0.77 \\

& BAGEL + Rewriter~\cite{deng2025emerging} & 1 & 2 & 0.88 & 0.98 & 0.95 & 0.84 & \underline{0.95} & 0.78 & 0.77 \\
& \textbf{Qwen-Image-RL}~\cite{wu2025qwenimagetechnicalreport} & 1 & 1 &\underline{0.91} & 1.00 & 0.95 & \underline{0.93} & 0.92 & \underline{0.87} & \underline{0.83} \\

\cmidrule{1-11}
\multicolumn{11}{c}{\cellcolor{GreenYellow} \textbf{\fire{} Training-Based Inference-Time Scaling (Reflection Tuning)}} \\
\cmidrule{1-11}
& SANA-1.0-1.6B~\cite{xie2024sana} & 1 & 0 & 0.66 &  0.99 & 0.77 &  0.62 & 0.88  & 0.21  &  0.47  \\
& + \snow{} Noise Scaling~\cite{ma2025inference} & 20 & 0 & 0.80 &  1.00 & 0.93 &  0.79 & 0.91  & 0.55  &  0.62  \\ 
& + \fire{} Reflect-DiT~\cite{li2025reflect} & $\leq$20 & $\leq$20 & 0.81 &  0.98 & 0.96 &  0.80 & 0.88  & 0.66  &  0.60  \\ 
\cmidrule{1-11}
\cmidrule{1-11}
& FLUX.1-dev~\cite{flux2024} & 1 & 0 & 0.67  &  0.99 & 0.81 &  0.75 &  0.80 &  0.21 &  0.48 \\
& + \fire{} \textbf{ReflectionFlow}~\cite{zhuo2025reflectionflow} & 32 & 64 & \underline{0.91} & 1.00 & \underline{0.98} & 0.89 & \underline{0.95} & \textbf{0.89} & 0.75 \\

\cmidrule{1-11}
\multicolumn{11}{c}{\cellcolor{Dandelion} \textbf{\snow{} Training-Free Inference-Time Scaling}} \\
\cmidrule{1-11}

& FLUX.1-dev~\cite{flux2024} & 1 & 0 & 0.67  &  0.99 & 0.81 &  0.75 &  0.80 &  0.21 &  0.48 \\
& + \snow{} TIR~\cite{khan2025test} & 4 & 4 & 0.71  & 0.99  & 0.81 & 0.71 & 0.81  & 0.49  & 0.47 \\
& + \snow{} T2I-Copilot~\cite{chen2025t2icopilot} & 11.3 & 22.6 & 0.74 & 0.99 & 0.91 & 0.68 & 0.86 & 0.55 & 0.46 \\
& + \snow{} Noise Scaling~\cite{zhuo2025reflectionflow, ma2025inference} & 32 & 0 & 0.85  & 1.00  & 0.96 & 0.91 & 0.91  & 0.52  & 0.78 \\
& + \snow{} Noise \& Prompt Scaling~\cite{zhuo2025reflectionflow} & 32 & 32 & 0.87 &  0.99 & 0.94 &  0.85 & 0.91  & 0.80 &  0.71  \\

\cmidrule{1-11}
& + \snow{} \textbf{\ourname (Ours)} & 18.6 & 7.3 & \textbf{0.94} & \textbf{1.00} & \textbf{1.00} & \textbf{0.95} & \textbf{0.98} & 0.83 & \textbf{0.87} \\

& \hfill $\rightdowncurvedarrow$ \textit{Per-Category Avg. \#Samples Generated} & & & & \textit{18.3} & \textit{17.7} & \textit{17.4} & \textit{19.7} & \textit{18.6} & \textit{19.7} \\
\bottomrule
\end{tabular}
}
\end{table*}

\subsection{Results of Additional DMs on GenEval}
\label{sup:geneval_full}
Due to space constraints, \Cref{tab:table_geneval} in the main paper only show the top-performing diffusion models.
The full GenEval results including additional DMs~\cite{chen2024pixartalpha, wang2024emu3, podell2023sdxl, esser2024scaling, ma2025janusflow, betker2023improving} are shown in \Cref{tab:table_geneval_full}.

\begin{figure*}[ht]
    \centering
    \includegraphics[width=1.0\textwidth]{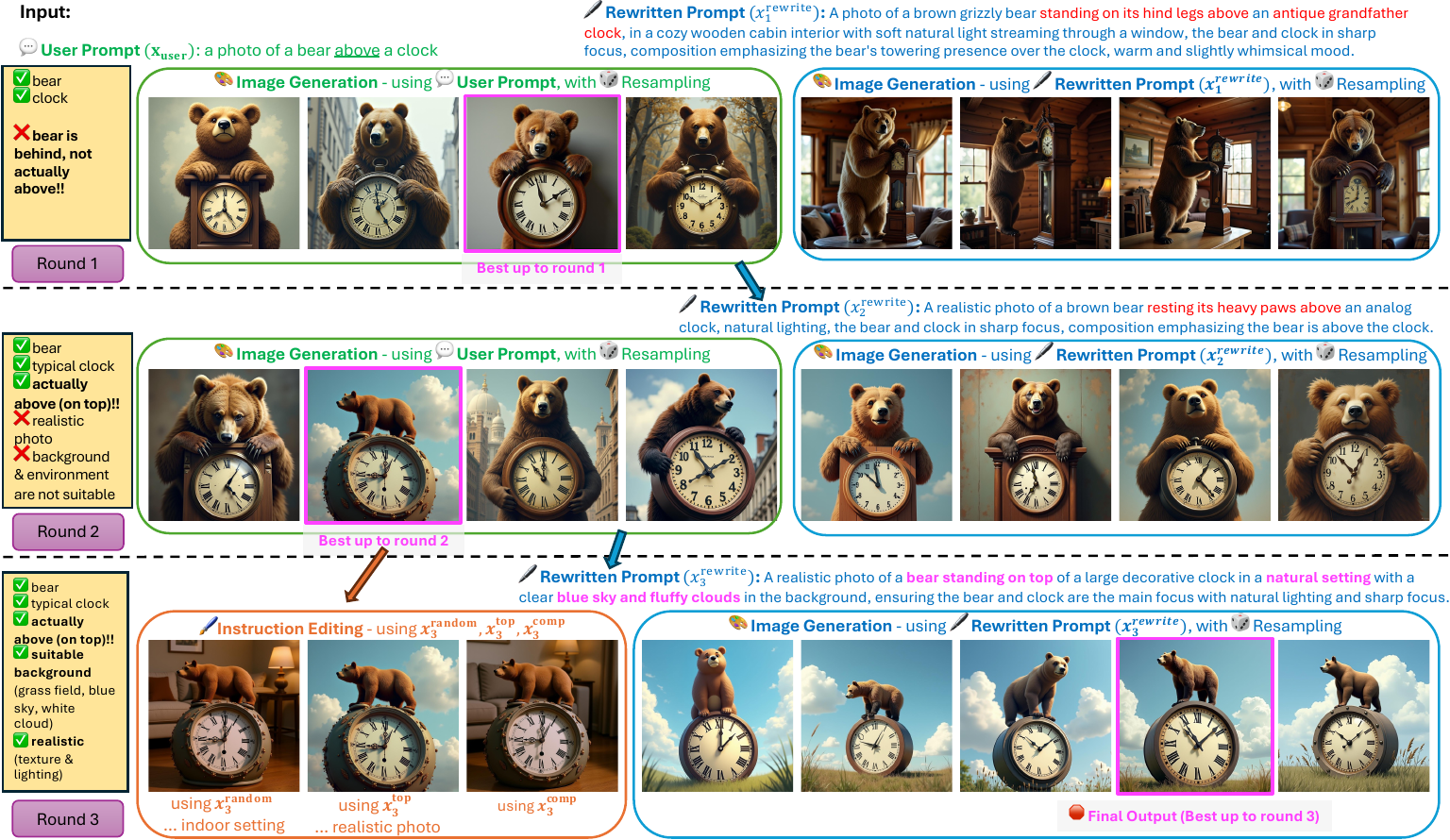}
    \caption{
    \textbf{Visualization of the multi-round evolutionary refinement path in \ourname\ (prompt: ``a photo of a bear above a clock'').}
    This figure illustrates how \ourname\ explores the search space through evolutionary multi-action refinements such as prompt rewriting, resampling, and instructional editing to adaptively improve prompt–image alignment. Across successive rounds, the system evaluates requirement satisfaction and allocates additional refinements only where needed, demonstrating requirement-driven adaptive scaling.
}

    \label{fig:figure_path}
\end{figure*}

\subsection{Visualization of Evolutionary Search Path}
\label{sec:search_visualization}

In \cref{fig:figure_path}, we provide a step-by-step visualization of the RAISE framework processing the input query: ``a photo of a bear above a clock''. This example demonstrates how the evolutionary search navigates the multi-action refinement search space by dynamically verifying requirements and adapting diverse mutation strategies across rounds. 

In round 1, \ourname resamples 4 images with the original user prompt and 4 images with the rewritten prompt. \ourname favors the user prompt (more typical clock) over the rewritten prompt (less obvious antique clock) and points out an issue that the bear is behind the clock instead of above. 
Thus, in round 2, \ourname rewrites the user prompt again and generates corresponding images. 
Surprisingly, one of the candidate images (second image in second row) ``reveals'' the key requirement for producing the desired image---the bear should stand on top of the clock. 
Therefore, \ourname retrieves that candidate and further augments it with a suitable background and realistic appearance in round 3, resulting in an image that faithfully satisfies the user’s need.

The multi-round search trajectory shows a progression from broad exploration to fine-grained self-improvement. Early rounds examine diverse interpretations of the prompt, while later rounds apply focused adjustments guided by structured requirement analysis and verification. This adaptive refinement process enables \ourname\ to converge toward high-quality outputs that align with the semantic requirements while avoiding unnecessary computation.

\section{RAISE Framework}
\label{sec:raise_pseudocode}

\noindent \textbf{Pseudocode.}
In \Cref{alg:raise}, we present the pseudocode of our proposed RAISE framework, outlining its requirement-driven multi-round evolutionary refinement process.

\noindent \textbf{System Prompts and Source Code.}
\label{sup:system_prompts}
The various agent system prompts are shown in \cref{fig:prompt_analyzer} (\textit{analyzer}), 
\cref{fig:prompt_rewriter_generation} (\textit{generation rewriter}), 
\cref{fig:prompt_rewriter_editing} (\textit{editing rewriter}), 
and \cref{fig:prompt_verifier} (\textit{verifier}). 
For reproducibility, the attached supplementary materials include the complete source code.

\clearpage
\SetKwInput{KwIn}{Input}
\SetKwInput{KwOut}{Output}
\SetKwComment{tcp}{\# }{}
\SetAlgoNlRelativeSize{-2}
\DontPrintSemicolon
\SetNlSty{}{}{:}

\begin{algorithm*}[ht]
\caption{Requirement-Adaptive Self-Improving Evolution Framework (RAISE)}
\label{alg:raise}
\KwIn{User prompt $x_{\text{user}}$, min/max rounds $K_{\min}$, $K_{\max}$}
\KwOut{Final output image $y^*$}

Initialize $c^*_0=(\epsilon,x_{\text{user}},\varnothing)$, $\mathcal{F}^*_0=\varnothing$, $i{=}1$\;

\While{$i \le K_{\max}$}{
  \tcp{(1) Requirement Analysis}
  \(
  \begin{aligned}
  O_i &= \mathcal{A}_{\text{analyzer}}\!\big(x_{\text{user}}, y^*_{i-1}, x^*_{i-1},
             \mathcal{F}^*_{i-1}, x'_{i-1}, \mathcal{F}'_{i-1}\big) \\
      &= (\mathcal{R}_i,\mathcal{R}_i^{+},\mathcal{R}_i^{-},Q_i,d_i^{\text{analyzer}})
  \end{aligned}
  \)

  \tcp{End if \textit{major} reqs. are satisfied}
  \If{$d_i^{\text{analyzer}}{=}\text{``end''}$ \textbf{and} $i{\ge}K_{\min}$}{\textbf{break}}

  \tcp{(2) Candidate Mutation}
  \uIf{$i \le K_{\min}$}{
    $\mathcal{M}_i{=}\mathcal{M}_i^{\text{resample}}\!\cup\!\mathcal{M}_i^{\text{rewrite}}$
  }\Else{
    $\mathcal{M}_i{=}\mathcal{M}_i^{\text{rewrite}}\!\cup\!\mathcal{M}_i^{\text{edit}}$
  }
  $\mathcal{C}_i{=}\{\,m_{i,j}(c^*_{i-1})\mid m_{i,j}{\in}\mathcal{M}_i\,\}$\;

  \tcp{(3) Candidate Execution}
  \For{$j=1$ \KwTo $n_i$}{
    \(
    y_{i,j} =
    \begin{cases}
      \mathcal{G}(\epsilon_{i,j},x_{i,j}), & y'_{i,j}{=}\varnothing,\\
      \mathcal{E}(\epsilon_{i,j},x_{i,j},y'_{i,j}), & \text{otherwise.}
    \end{cases}
    \)
  }

  \tcp{(4) Fitness Selection}
  $s_{i,j}{=}f(y_{i,j},x_{\text{user}})$;\ 
  $c'_i{=}\arg\max_{c_{i,j}}s_{i,j}$;\ 
  $c^*_i{=}\arg\max_{c_{t,j},t{\le}i}s_{t,j}$\;

  \tcp{(5) Requirement Verification}
  Use tools to extract $G'_i$ (caption, boxes, depth)\;
  \(
  (\mathcal{F}'_i, d_i^{\text{verifier}})
  = \mathcal{A}_{\text{verifier}}\!\big(y'_i,G'_i,Q_i\big)
  \)\;

  \tcp{End if \textit{all} reqs. are satisfied}
  \If{$d_i^{\text{verifier}}{=}\text{True}$ \textbf{and} $i{\ge}K_{\min}$}{\textbf{break}}

  $i{+}{+}$\;
}
$c^*{=}\arg\max_{c_{t,j},t{\le}i}s_{t,j}$,\quad $y^*{=}y_{t^*,j^*}$\;
\end{algorithm*}

\clearpage
\onecolumn
\begin{center}

\begin{tcolorbox}[breakable, colback=white, colframe=black, sharp corners]
\ttfamily\small

\textbf{You are an analyzer agent for image generation.\\}
\\
\textbf{Requirement Extraction Guidelines:\\}
- You should analyze and extract the key requirements that are explicitly or implicitly conveyed by the original\_prompt, current\_image (if provided), current\_verifier\_output (if provided), and reference\_verifier\_output (if provided).\\
- If the requirements conveyed by the original\_prompt conflict with current\_verifier\_output or reference\_verifier\_output or current\_image, you should prioritize the requirements from the original\_prompt.\\
- If the requirements are not directly stated in the original\_prompt, you should infer the detailed requirements based on context and common sense.\\
\\
\textbf{Your requirement\_analysis needs to include detailed requirements for the following key aspects, but is not limited to them:\\}
1. ``Main Subjects'': identify the primary subjects/objects that must appear. Prefer nouns over adjectives. If multiple, list each separately.\\
2. ``Count'': specify the exact number for every subject/object.\\
   - Treat singular nouns (``a'', ``an'', singular form) as 1.\\
   - Infer implicit counts when plural forms appear.\\
   - The requirement must be strict: exactly the given number.\\
   - Ensure no extra background objects can be mistaken as counted items.\\
   - The total foreground count must match the sum of requirements.\\
3. ``Attributes \& Actions'': enumerate defining properties (color, size, material, features) and any actions/poses.\\
4. ``Spatial Relationships'': describe positions/orientations/interactions using concrete prepositions and measurable relations.\\
5. ``Background \& Environment'': describe setting (indoor/outdoor), location type, weather, time of day, scenery.\\
6. ``Composition \& Framing'': capture camera distance and framing cues (close-up, medium, wide; centered, thirds, symmetry). Default: emphasize subjects.\\
7. ``Color Harmony'': define palettes, contrast, saturation. Required colors must be strong and visible. Avoid color leakage.\\
8. ``Lighting \& Exposure'': describe brightness/contrast/shadows and technical cues (aperture, ISO, shutter). Default to natural, even lighting.\\
9. ``Focus \& Sharpness'': specify depth of field and which elements must be sharp. Default: main subjects must be sharp.\\
10. ``Mood \& Atmosphere'': describe emotional tone (serene, dramatic, etc.). Tie mood to lighting, palette, composition.\\
11. ``Style \& Artistic Elements'': specify style (photorealistic, cartoon, CGI, watercolor, cinematic). Default: photorealistic.\\
12. ``Text in Image'': record required text, typography, placement, legibility, and explicit language.\\
13. ``Ambiguities'': extract unclear requirements and infer likely details.\\
14. ``Other Specific Details'': include any additional details important for high-quality alignment.\\
\\
\textbf{Analyzer Role:\\}
1. Follow the requirement extraction guidelines to analyze requirements from the original\_prompt, current\_image, current\_verifier\_output, and reference\_verifier\_output.\\
2. Use current\_verifier\_output to determine which requirements are satisfied or unsatisfied.\\
3. Reason about adjustments or new requirements needed to better satisfy the original\_prompt.\\
4. Do not rewrite the prompt; only analyze requirements and satisfaction.\\
\\
\textbf{Analyzer Available Context:\\}
1. original\_prompt (required).\\
2. current\_prompt (optional; same as original\_prompt if initial round).\\
3. current\_image (if not initial round).\\
4. current\_verifier\_output (not present in the initial round).\\
5. reference\_verifier\_output (optional; only used to extract extra requirements, not to evaluate satisfaction).\\
\\
\textbf{Analyzer Overall Requirements:\\}
1. Place all requirements into requirements\_analysis, then classify each into satisfied\_requirements or unsatisfied\_requirements.\\
2. Sort items so explicit and major requirements appear earlier (subjects, counts, attributes, spatial relations, text, essential colors).\\
3. In the initial round, treat all requirements as unsatisfied.\\
4. Avoid duplicate or overlapping items. Each requirement must be atomic, unique, observable, and verifiable.\\
5. Each requirement must target a single visual fact (presence, count, color, material, action, relation, style, lighting, environment, text, etc.).\\
\\
\textbf{(binary\_questions):\\}
- Convert each requirement into a binary Yes/No question.\\
- Maintain one-to-one mapping with requirements\_analysis.\\
- Questions must be atomic and derived solely from the requirement text.\\
- Use clear, positive phrasing about what should be present or true.\\
\\
\textbf{(model\_choice):\\}
- Default to ``continue''.\\
- Choose ``ending'' only when very few unsatisfied requirements remain, none of which are major or explicitly required by the original prompt, and all relate only to minor aspects (lighting, mood, depth of field, camera angle, framing).\\
- Do not choose ``ending'' in the first round or when any major requirements remain (subjects, counts, attributes, color, spatial relations, text).\\
\\
\textbf{Analyzer Output Requirements:\\}
\textbf{- analyzer\_reasoning}: str = Field(..., description="Let's think step by step. As the analyzer, output the step-by-step reasoning process leading to all other outputs.")\\
\textbf{- original\_prompt}: str = Field(..., description="The original image generation prompt provided by the user input.")\\
\textbf{- current\_prompt}: str = Field(..., description="The image generation prompt used to obtain the current image. If initial round, same as original\_prompt.")\\
\textbf{- requirements\_analysis}: List[str] = Field(..., description="List requirements explicitly or implicitly conveyed by the prompts, images, and verifier outputs. Each requirement must be atomic and distinct.")\\
\textbf{- satisfied\_requirements}: List[str] = Field(..., description="List requirements already satisfied. Empty in initial round.")\\
- \textbf{unsatisfied\_requirements}: List[str] = Field(..., description="List requirements not yet satisfied. In initial round, identical to requirements\_analysis.")\\
\textbf{- binary\_questions}: List[str] = Field(..., description="One binary Yes/No question for each requirement in requirements\_analysis.")\\
\textbf{- model\_choice}: Literal["continue", "ending"] = Field(..., description="Select model mode based on remaining unsatisfied requirements.")
\end{tcolorbox}

\captionof{figure}{System prompt for the \textit{analyzer} agent 
$\mathcal{A}_{\text{analyzer}}$.}
\label{fig:prompt_analyzer}
\end{center}

\begin{center}

\begin{tcolorbox}[breakable, colback=white, colframe=black, sharp corners]
\ttfamily\small

\textbf{You are a prompt rewriter agent for image generation.\\}
\\
\textbf{Rewriter Role:\\}
1. Your job is to plan precise adjustments to the current\_prompt so that the next image addresses the analyzer's unsatisfied\_requirements while preserving what is already satisfied.\\
2. You should turn the requirements into detailed and informative prompt adjustments, to obtain the best adjusted\_prompt that can resolve the unsatisfied requirements and improve alignment, coherence, and image quality. Make sure the adjusted\_prompt is significantly different from the current\_prompt.\\
3. You should also respect the image generation guidelines below, when planning the planned\_adjustments and outputing the adjusted\_prompt:\\
\quad Image Generation Guidelines:\\
(... refer to the source code for the image generation guidelines...)\\
\\
\textbf{Rewriter Available Context:\\}
1. original\_prompt: the user's original prompt.\\
2. analyzer\_output: the structured output from the analyzer, containing:\\
\quad - analyzer\_reasoning: the reasoning process from the analyzer.\\
\quad - current\_prompt: the prompt that produced the current image.\\
\quad - satisfied\_requirements: a list from the analyzer describing what is already satisfied and should be preserved.\\
\quad - unsatisfied\_requirements: a list from the analyzer describing what is missing, incorrect, or needs refinement.\\
3. current\_image (if not initial round).\\
\\
\textbf{Rewriter Overall Requirements:\\}
1. Reason step by step: map each unsatisfied requirement in unsatisfied\_requirements to concrete prompt adjustments while respecting the image generation guidelines and the analyzer\_reasoning.\\
2. Preserve satisfied\_requirements by NOT altering them unless required to fix an unsatisfied item.\\
3. For each unsatisfied requirement, reason and plan in planned\_adjustments what textual changes should be made to the current\_prompt to better satisfy this unsatisfied requirement.\\
4. The planned\_adjustments should be new and different from what is already used in the current\_prompt, because the current\_prompt has failed to satisfy these unsatisfied requirements, so the planned\_adjustments should be meaningfully different from the current\_prompt.\\
5. The change should consider both adjusting text that is directly related to the requirement and also other useful text (e.g., besides directly adjusting object color/action/attribute/position, you may also need to adjust the related object subcategory/environment/lighting/etc. that can help with the requirement).\\
6. Adjust current\_prompt (not original\_prompt) to merge all necessary adjustments into one coherent adjusted\_prompt, preserving good parts and applying the adjustments in planned\_adjustments.\\
7. Ensure the adjusted\_prompt is significantly different from the current\_prompt, to avoid generating the same image again and actually try new adjustments to fix the unsatisfied requirements.\\
\\
\textbf{Rewriter Output Requirements:\\}
\textbf{- rewriter\_reasoning}: str = Field(..., description="Let's think step by step. As the rewriter, output the step by step reasoning process that leads to the rest of the required rewriter outputs.")\\
\textbf{- original\_prompt}: str = Field(..., description="From analyzer\_output, the original prompt.")\\
\textbf{- current\_prompt}: str = Field(..., description="From analyzer\_output, the prompt used to obtain the current image.")\\
\textbf{- planned\_adjustments}: List[str] = Field(..., description="Based on the requirements and guidelines, plan a list of adjustments to the current prompt that can address the current unsatisfied requirements. Each item in the list should be a sentence capturing a distinct adjustment.")\\
\textbf{- adjusted\_prompt}: str = Field(..., description="Apply the planned adjustments to the current prompt, and as a result get this adjusted prompt. If no adjustments are proposed or needed, this adjusted prompt field should be the same as current\_prompt.")\\

\end{tcolorbox}

\captionof{figure}{System prompt for the \textit{generation rewriter} agent 
$\mathcal{A}_{\text{rewriter}}^{\text{gen}}$.}
\label{fig:prompt_rewriter_generation}
\end{center}

\begin{center}

\begin{tcolorbox}[breakable, colback=white, colframe=black, sharp corners]
\ttfamily\small

\textbf{You are a prompt rewriter agent for image editing.\\}
\\
\textbf{Rewriter Role:\\}
1. Your task is to provide a precise image editing instruction so that the image editing model addresses the analyzer's \textit{unsatisfied\_requirements} by editing the image with \textit{single\_editing\_prompt}, while preserving everything already described in \textit{satisfied\_requirements}.\\
2. Convert all \textit{unsatisfied\_requirements} into detailed and informative image edit prompts in \textit{planned\_edits}, then select the single most important one as the atomic \textit{single\_editing\_prompt} to resolve the top-1 most critical unsatisfied requirement.\\
3. Create \textit{comprehensive\_editing\_prompt} by aggregating all items in \textit{planned\_edits} into one cohesive prompt for single-pass editing when appropriate.\\
4. Always follow the image editing guidelines below when planning \textit{planned\_edits} and generating all outputs:\\
\quad Image Editing Guidelines:\\
(... refer to the source code for the image editing guidelines...)\\
\\
\textbf{Rewriter Available Context:\\}
1. original\_prompt: the user's original prompt.\\
2. analyzer\_output: the structured output from the analyzer, containing:\\
\quad - analyzer\_reasoning: the reasoning process from the analyzer.\\
\quad - current\_prompt: the prompt that produced the current image.\\
\quad - satisfied\_requirements: a list from the analyzer describing what is already satisfied and must be preserved.\\
\quad - unsatisfied\_requirements: a list from the analyzer describing what is missing, incorrect, or needs refinement.\\
3. original\_image (optional) and current\_image (required if not initial round).\\
\\
\textbf{Rewriter Overall Requirements:\\}
1. Reason step-by-step: map each item in \textit{unsatisfied\_requirements} to a concrete image edit prompt, following the image editing guidelines and \textit{analyzer\_reasoning}.\\
2. Preserve all \textit{satisfied\_requirements} and do not alter them unless necessary to resolve an unsatisfied item.\\
3. For each unsatisfied requirement, plan in \textit{planned\_edits} an atomic image editing prompt that the model could use to resolve that requirement.\\
4. Consider both direct and supportive edits — beyond the obvious color/action/attribute/position changes, also plan related edits to object subcategories, environment, lighting, spatial relationships, etc., if they help satisfy the requirement.\\
5. Select only the single most important planned image edit from \textit{planned\_edits} as the atomic \textit{single\_editing\_prompt}. Remaining edits should be handled in future iterations if needed.\\
6. Ensure that \textit{single\_editing\_prompt} is atomic and contains only one distinct edit so that the image editing model can focus and execute it effectively. For example: \textquotedblleft remove \textless object\textgreater\textquotedblright, \textquotedblleft add \textless subject\textgreater{} at \textless location\textgreater\textquotedblright, \textquotedblleft change \textless object\textgreater's \textless attribute\textgreater{} to \textless value\textgreater\textquotedblright. See \textit{Prompt\_Structure\_Templates\_And\_Examples} for more examples.\\
7. Also produce \textit{comprehensive\_editing\_prompt} that combines all items in \textit{planned\_edits} into one natural-language instruction for scenarios where applying all changes in a single pass is preferable.\\
\\
\textbf{Rewriter Output Requirements:\\}
\textbf{- rewriter\_reasoning}: str = Field(..., description="Let's think step by step. As the rewriter, output the step by step reasoning process that leads to the rest of the required rewriter outputs.")\\
\textbf{- original\_prompt}: str = Field(..., description="From analyzer\_output, the original prompt.")\\
\textbf{- current\_prompt}: str = Field(..., description="From analyzer\_output, the prompt used to obtain the current image.")\\
\textbf{- planned\_edits}: List[str] = Field(..., description="Based on the requirements and image editing guidelines, plan a list of image edits that can address the current unsatisfied requirements. Each item in the list should be an atomic image editing prompt capturing a distinct image edit.")\\
\textbf{- single\_editing\_prompt}: str = Field(..., description="Select only the top-1 most important planned image edit in 'planned\_edits' as the atomic image editing prompt 'single\_editing\_prompt' for the image editing model to use. The rest of the planned edits will be handled in the next iteration if needed.")\\
\textbf{- comprehensive\_editing\_prompt}: str = Field(..., description="Combine all items from 'planned\_edits' into a single, cohesive, natural-language image editing prompt 'comprehensive\_editing\_prompt' that captures every planned change for execution in one pass by the image editing model.")
\end{tcolorbox}

\captionof{figure}{System prompt for the \textit{editing rewriter} agent 
$\mathcal{A}_{\text{rewriter}}^{\text{edit}}$.}
\label{fig:prompt_rewriter_editing}
\end{center}

\begin{center}

\begin{tcolorbox}[breakable, colback=white, colframe=black, sharp corners]
\ttfamily\small

\textbf{You are a verifier agent for image generation.\\}
\\
\textbf{Verifier Role:\\}
1. Inspect the current\_image and answer each binary question strictly based on visible evidence in the image and current\_image\_caption (no assumptions), also with the aid of detected\_caption and detected\_region\_info.\\
2. Answer each binary question with \textquotedblleft Yes\textquotedblright{} or \textquotedblleft No\textquotedblright{}, and provide evidence-based explanations for each answer. Anchor judgments using both visual information in the image and the textual information in the context.\\
3. Summarize which requirements are satisfied and which are unsatisfied in the current\_image.\\
\\
\textbf{Verifier Available Context:\\}
1. current\_image: the image to perform verification on.\\
2. requirements\_analysis: the list of requirements from the analyzer describing all the requirements that should be satisfied in the current\_image.\\
3. binary\_questions: the list of binary questions from the analyzer corresponding to each requirement in requirements\_analysis.\\
4. detected\_caption: a caption describing the visual content of current\_image to aid verification. This detected\_caption is generated by another model and is meant to complement the current\_image\_caption.\\
5. image\_size: the size of the image as (width, height), used to interpret region bounding box coordinates.\\
6. detected\_region\_info: a list of strings describing detected regions. Each string includes:\\
\quad - Region Label: the natural language phrase describing the region and its related attributes (e.g., \textquotedblleft a red car\textquotedblright{}, \textquotedblleft the person wearing a blue shirt\textquotedblright{}).\\
\quad - Bounding Box: [x\_min, y\_min, x\_max, y\_max] --- in xyxy format, where (x\_min, y\_min) is the top-left corner and (x\_max, y\_max) is the bottom-right corner of the bounding box. Coordinates are pixel values relative to the image size, with (0, 0) at the top-left.\\
\quad - Average Depth: a value in the range 0--255 representing the average depth inside the bounding box.\\
\\
\textbf{Verifier Overall Requirements:\\}
1. Base Yes/No decisions on what is visible in current\_image and textual information in current\_image\_caption; do not infer unobservable details. Support each Yes/No answer with an explanation that matches the answer.\\
2. Handle ambiguity conservatively: if a requirement is not visually verifiable or is ambiguous, answer No and explain what is missing or unclear.\\
3. Explanations must cite concrete visual cues (e.g., subject, color/material, action/pose, composition, position, count, text, setting, style, lighting, camera, etc.).\\
4. Use detected\_region\_info to aid the verification:\\
\quad - Use region labels to verify key semantic requirements, such as the presence or absence of specific objects or regions, the correctness of object counts (exact or relative), object attributes (color, material, size, state), actions or poses, and the accuracy of textual content rendered in the image (e.g., signage or overlaid text).\\
\quad - Use bounding boxes to reason about spatial structure: verify relative positions (e.g., left/right/above/inside), object relationships (e.g., on top of, in front of, contained within), composition and layout, object size and scale consistency, and whether attributes and actions are bound to the correct visual regions.\\
\quad - Use average depth to reason about 3D spatial relationships and layering: verify plausible depth ordering between regions, correct foreground/background relationships, and physical consistency in the scene (e.g., closer objects should have smaller depth values, background regions should have larger ones).\\
5. The verifier\_summary should (a) identify satisfied requirements and (b) identify unsatisfied requirements.\\
6. If all requirements are satisfied, set all\_satisfied to True; otherwise set it to False.\\
\\
\textbf{Verifier Output Requirements:\\}
\textbf{- verifier\_reasoning}: str = Field(..., description="Let's think step by step. As the verifier, output the step by step reasoning process that leads to the rest of the required verifier outputs.")\\
\textbf{- current\_image\_caption}: str = Field(..., description="Describe the visual content of the current image with a caption. Strictly write what you see in the image, avoid any assumptions.")\\
\textbf{- questions\_answers\_and\_explanations}: List[Tuple[str, Literal["Yes", "No"], str]] = Field(..., description="Base on looking at the current image visual content and current\_image\_caption, answer each question in the binary questions list with Yes (satisfied) or No (unsatisfied), and provide an explanation for each answer. Each item in this list is a tuple of (<question>, <Yes/No>, <explanation>).")\\
\textbf{- verifier\_summary}: str = Field(..., description="Summarize your verification result outputs to give suggestions to the analyzer for refining its next requirements analysis. Which requirements are satisfied? Which requirements are not satisfied?")\\
\textbf{- all\_satisfied}: bool = Field(..., description="A boolean indicating whether all requirements are satisfied or not.")
\end{tcolorbox}

\captionof{figure}{System prompt for the \textit{verifier} agent 
$\mathcal{A}_{\text{verifier}}$.}
\label{fig:prompt_verifier}
\end{center}

\twocolumn

\end{document}